%% file: bare_jrnl.tex
\documentclass[journal]{IEEEtran}
\ifCLASSINFOpdf
\else
\fi

\usepackage{graphicx}

\usepackage{amsmath}
\usepackage{multirow}
\usepackage{soul}
\usepackage{subfigure}

\setul{0.2ex}{0.15ex}

\usepackage{amssymb}
\usepackage{amsfonts}
\usepackage{booktabs}
\usepackage{graphicx}
\usepackage{bbding}
\usepackage{pifont}
\usepackage{amsmath}
\usepackage{wasysym}
\usepackage{makecell}
\usepackage{multicol}
\usepackage{multirow}
\usepackage{soul}
\usepackage{xcolor}
\usepackage{pifont}
\usepackage{subfigure}
\usepackage{subcaption}
\usepackage{lineno}
\usepackage[hyphenbreaks]{breakurl}
\usepackage{hyperref}
\usepackage{algorithm,algorithmic}
\usepackage{float}
\usepackage{placeins}

\begin{document}
\title{ResPanDiff: Diffusion Model for Pansharpening by Inferring Residual Inference}

\DeclareRobustCommand*{\authorrefmarknumber}[1]{%
    \raisebox{0pt}[0pt][0pt]{\textsuperscript{\footnotesize\ensuremath{#1}}}}

\author{
\IEEEauthorblockN{
ShiQi~Cao{\authorrefmarknumber{1}},
Shangqi~Deng{\IEEEauthorrefmark{1}\authorrefmarknumber{1}},
Shangqi~Deng{\authorrefmarknumber{1}},
}
}

\markboth{Journal of \LaTeX\ Class Files,~Vol.~13, No.~9, September~2014}%
{Shell \MakeLowercase{\textit{et al.}}: Bare Demo of IEEEtran.cls for Journals}



\maketitle

\begin{abstract}
\input{sections/abstract}
\end{abstract}


% \newcommand\blfootnote[1]{%
%   \begingroup
%   \renewcommand\thefootnote{}\footnote{#1}%
%   \addtocounter{footnote}{-1}%
%   \endgroup
% }

% \blfootnote{Working in progress.}

% \fi
\IEEEpeerreviewmaketitle

%=====================MAIN PART======================
\input{sections/intro}
\input{sections/related-works}

\input{sections/method}
\input{sections/exps}
\appendix \input{sections/appendix}

\appendices

% \section{}

% \section*{Acknowledgment}


\ifCLASSOPTIONcaptionsoff
  \newpage
\fi



\bibliographystyle{IEEEtran}
\bibliography{bibtex/bib/reference}

\end{document}

%% file: sections/abstract.tex
The implementation of diffusion-based pansharpening task is predominantly constrained by its slow inference speed, which results from numerous sampling steps. Despite the existing techniques aiming to accelerate sampling, they often compromise performance when fusing multi-source images. To ease this limitation, we introduce a novel and efficient diffusion model named Diffusion Model for Pansharpening by Inferring Residual Inference (ResPanDiff), which significantly reduces the number of diffusion steps without sacrificing the performance to tackle pansharpening task. In ResPanDiff, we innovatively propose a Markov chain that transits from noisy residuals to the residuals between the LRMS and HRMS images, thereby reducing the number of sampling steps and enhancing performance. Additionally, we design the latent space to help model extract more features at the encoding stage, Shallow Cond-Injection~(SC-I) to help model fetch cond-injected hidden features with higher dimensions, and loss functions to give a better guidance for the residual generation task. enabling the model to achieve superior performance in residual generation.
Furthermore, experimental evaluations on pansharpening datasets demonstrate that the proposed method achieves superior outcomes compared to recent state-of-the-art~(SOTA) techniques, requiring only 15 sampling steps, which reduces over $90\%$ step compared with the benchmark diffusion models. Our experiments also include thorough discussions and ablation studies to underscore the effectiveness of our approach. 

%% file: sections/intro.tex
\section{Introduction}
\begin{figure}[!t]
    \centering
    \includegraphics[width=1.0\linewidth]{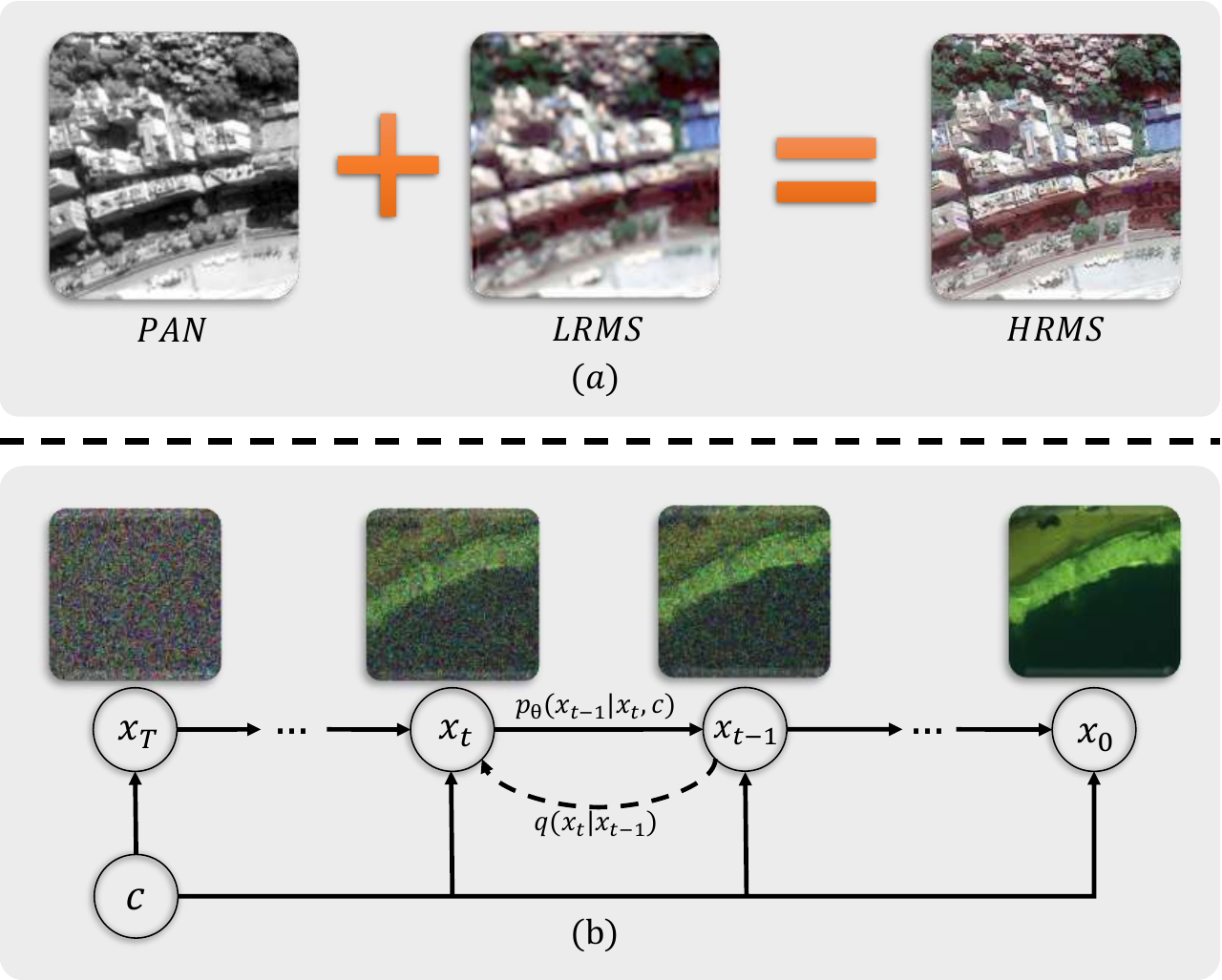}
    \caption{(a) Pansharpening involves fusing the PAN and LRMS images into an HRMS image. (b) The process of the current denoising diffusion model utilized in pansharpening. The $q_(x_t|x_{t-1})$, $p_\theta(x_{t-1}|x_t,c)$, and $c$ represent the noise-adding forward process, the denoising backward process, and the condition, respectively.}
    \label{fig: introduce of pansharpening}
\end{figure}

\IEEEPARstart{P}Pansharpening, as a fundamental issue in multi-source image fusion (MSIF), has received extensive research attention in recent years. This technology aims to fuse low spatial resolution multi-spectral images (LRMS) with high spatial resolution panchromatic images (PAN) to generate high spatial resolution multi-spectral images (HRMS), as illustrated in Fig.~\ref{fig: introduce of pansharpening}. 

Therefore, the extraction of multispectral image background information and high-resolution panchromatic image spatial information is crucial for the effectiveness of panchromatic sharpening. Effectively extracting these latent features has become a key focus of research. 
Currently, based on different feature extraction methods, pansharpening techniques can be categorized into four main categories.
(1) component substitution (CS) methods~\cite{kwarteng1989extracting, laben2000process, aiazzi2007improving, qu2017hyperspectral}, (2) multiresolution analysis (MRA) methods~\cite{otazu2005introduction, liu2000smoothing, aiazzi2006mtf, vivone2013contrast}, (3) variational optimized (VO) techniques~\cite{he2014new, moeller2009variational, wang2018high, wu2023lrtcfpan}, (4) deep learning (DL) approaches~\cite{pnn, yang2017pannet, deng2020detail, 10504844}. Categories 1 to 3 represent traditional pansharpening approaches, and the fourth approach is most related to our work.

CS-based pansharpening methods enhance spatial information by projecting LRMS images into a new feature space, replacing their structural components with those from the PAN image. 
% Techniques like PCA, IHS transformation, and the Gram-Schmidt process differ in projection rules. 
While CS methods improve spatial quality, they often introduce spectral distortion. MRA-based pansharpening methods aim to enhance spatial resolution by infusing high-frequency details from the PAN image into LRMS images via multi-scale decomposition. 
% Techniques like wavelet and pyramid Laplacian transforms underpin methods such as MTF-GLP, ATWT, and AWLP. 
These approaches typically achieve high spectral fidelity but may compromise spatial quality. VO-based pansharpening methods approach the task as an ill-posed inverse problem, framed as a variational optimization process with a probabilistic image model. These methods involve two key steps: constructing an objective function and solving it using iterative optimization algorithms. The objective typically includes a fidelity term and regularization terms based on prior knowledge. 
% Notable methods include P+XS, Bayesian approaches, and sparse representation techniques.
While VO methods offer a good balance between spatial and spectral quality, they are computationally intensive and may underperform if their assumptions do not align with the fusion scenario.

Due to the outstanding feature extraction and aggregation capabilities of deep neural networks, DL-based methods have become a significant research trend in recent years. Convolutional Neural Networks (CNN)~\cite{resnet, unet} have garnered considerable attention for their impressive performance in pansharpening tasks. Various CNN-based methods have been proposed, including the first DL-based pansharpening approach, PNN~\cite{pnn}, PanNet~\cite{yang2017pannet}, TFNet~\cite{ResTFNet}. However, CNN-based methods may suffer from feature smoothing caused by vanilla convolutions, leading to suboptimal fusion results at boundaries.

Recent years, diffusion model has received much attention in multiple fields and has had awesome success, such as conditional image generation~\cite{ho2020denoising, karras2022elucidating}, image-to-image translation~\cite{saharia2022palette}, image super-resolution~\cite{2021arXiv210407636S, yue2024resshift} and pansharpening~\cite{10136205, cao2024diffusion}, also shown in Fig.~\ref{fig: introduce of pansharpening}. All the examples demonstrate the powerful generative capabilities of diffusion models. Owing to the structured forward and reverse processes, diffusion model offers a more stable training process and avoid issues like model collapse that are commonly seen in GAN-based models. Specifically, in the forward process, diffusion model builds up a Markov chain to gradually add noise to the data until it becomes pre-specified prior distribution. In the reverse process, it samples a noise map from the prior distribution and denoises it according to the reverse path of the Markov chain with large sampling steps. Although some techniques~\cite{ddim,improved_ddpm,dpm_solver} have been proposed to accelerate the sampling steps, these inevitably sacrifice the performance of the model. What's more, these methods may not be suitable for pansharpening since the reverse process starts with noise, which does not make full use of the information in LRMS and PAN images. It is reasonable to build up a new Markov chain that starts with LRMS and PAN. In addition, for the SR task, generating the residual between High-Resolution~(HR) and Low-resolution~(LR) images and adding it with LR image to get the target image is better than generating the HR image directly. In Sec.~\ref{sec: ResPanDiff}, we elucidate and validate the efficiency and importance of residual learning in the pansharpening task.

In response to the aforementioned shortcomings, we propose a novel diffusion model that starts from a prior distribution based on the residual between the LRMS image and HRMS image, making it possible that the HRMS image can be recovered from adding the LRMS image with its residual, which comes from a shorter Markov chain starting from a noisy residual whose distribution of its latent space approximates LRMS rather than a gaussian white noise. This design can decrease the number of diffusion sampling steps, improving inference efficiency. Regarding the generative model section, to enhance model performance, we design the Latent Space, Shallow Cond-Injection, and Loss Function tailored to the characteristics of residual generation and the MSIF task. These adaptations help the model better handle residual generation and the MSIF task. 
% We achieve this by carefully design a model that extracting features in both LRMS and PAN image to generate the residual at every $t$ step, Effectively increasing the generating result. 
To sum up, the main contributions of this work are as follows:
\begin{enumerate}
    \item We present an innovative diffusion model named ResPanDiff tailored for image fusion, which leverages a Markov chain to generate samples that accurately match the residuals. This approach effectively addresses the challenge of slow sampling speeds during the inference stage. ResPanDiff not only accelerates the inference process significantly but also ensures the preservation of high image quality.

    \item We have developed several methods for the generative model to help it achieve MSIF and residual generation tasks better. The generation model predicts the target residual by feeding both the latent space and the condition involving LRMS and PAN. This design effectively helps the model to extract more efficient features. With the designed loss function, the model receives more guided gradients as it approaches optimal performance, enabling it to achieve superior results.

    \item We conduct numerous experiments, demonstrating that ResPanDiff achieves state-of-the-art (SOTA) performance on the pansharpening task on three widely used pansharpening datasets. Comprehensive discussions and ablation studies further validate the effectiveness of the proposed method.
    
\end{enumerate}

%% file: sections/related-works.tex
\section{Related works}
\label{sec:related}
This section mainly introduces diffusion methods that are mostly related to our work.

\subsection{Diffusion models for pansharpening task}
\begin{figure}[!t]
    \centering
    \includegraphics[width=0.7\linewidth]{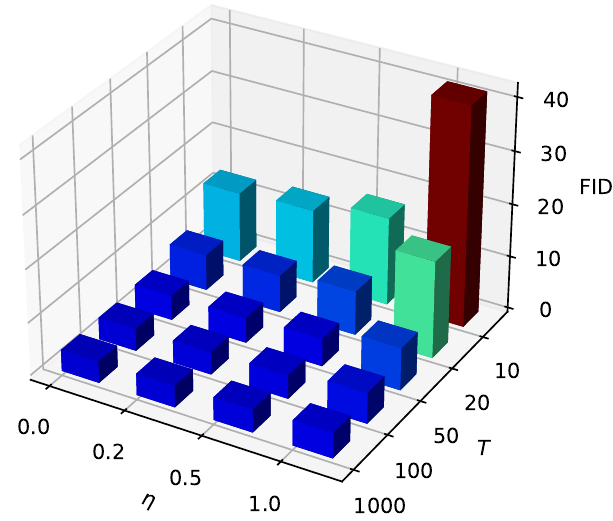}
    \caption{The score of Frechet Inception Distance~(FID~\cite{heusel2018ganstrainedtimescaleupdate}), on the CIFAR10 dataset comes from DDIM~\cite{ddim}, where $\eta$ is a hyperparameter that is directly controlled (it includes an original DDPM generative process when $\eta$ = 1 and DDIM when $\eta$ = 0) and $T$ represents the total timesteps. We can find that the more timesteps DDIM reduces, the worse its performance becomes.}
    \label{fig: DDIM_FID}
\end{figure}

Unlike traditional image generation or image super-resolution tasks, pansharpening receives both LRMS and PAN images to generate HRMS. PanDiff~\cite{10136205} treats the LRMS and PAN as cond to guide the model to predict the target HRMS. DDIF~\cite{cao2024diffusion} further imports disentangled modulations as conditions and proposes two modulation modules to disentangle style information and frequency components from different domain conditions, which helps the model suit the pansharpening task. However, limited research is dedicated to these diffusion models for the pansharpening task. The giant time steps result in lengthy sampling times. Even though DDIF modified the sampling process of the DDIM ODE sampler~\cite{ddim} to solve the problem of low sampling speed, it inevitably sacrifices the performance of the model, which is shown in Fig.~\ref{fig: DDIM_FID}.

\subsection{Methods on accelerating diffusion sampling speed}
Several methods have been proposed to accelerate the sampling speed. Denoising DDIMs~\cite{ddim} introduces a more efficient class of iterative implicit probabilistic models, retaining the same training procedure as DDPMs~\cite{ho2020denoising}. DDIMs generalize DDPMs by leveraging a class of non-Markovian diffusion processes, which can correspond to generative processes. This results in implicit models that produce high-quality samples at a significantly faster rate. ResShift~\cite{yue2024resshift} utilizes residual shifting to enhance image quality. Instead of reconstructing an image from Gaussian white noise, the model incrementally adjusts residuals or differences between the low-resolution input and its high-resolution target. ResShift iteratively refines the image by shifting its residuals towards a high-resolution output using a learned diffusion process conditioned on the input image, resulting in faster sampling steps. 
However, the aforementioned methods are either designed for image super-resolution or image generation; it is not completely suitable for pansharpening tasks after all. This raises a question: \textit{how and if we can accelerate the diffusion model in the pansharpening task by utilizing an idea similar to this method in an end-to-end manner?}

%% file: sections/method.tex
\section{Methods}
\label{sec:method}
\begin{figure*}[!ht]
\centering
    \includegraphics[width=0.9\linewidth]{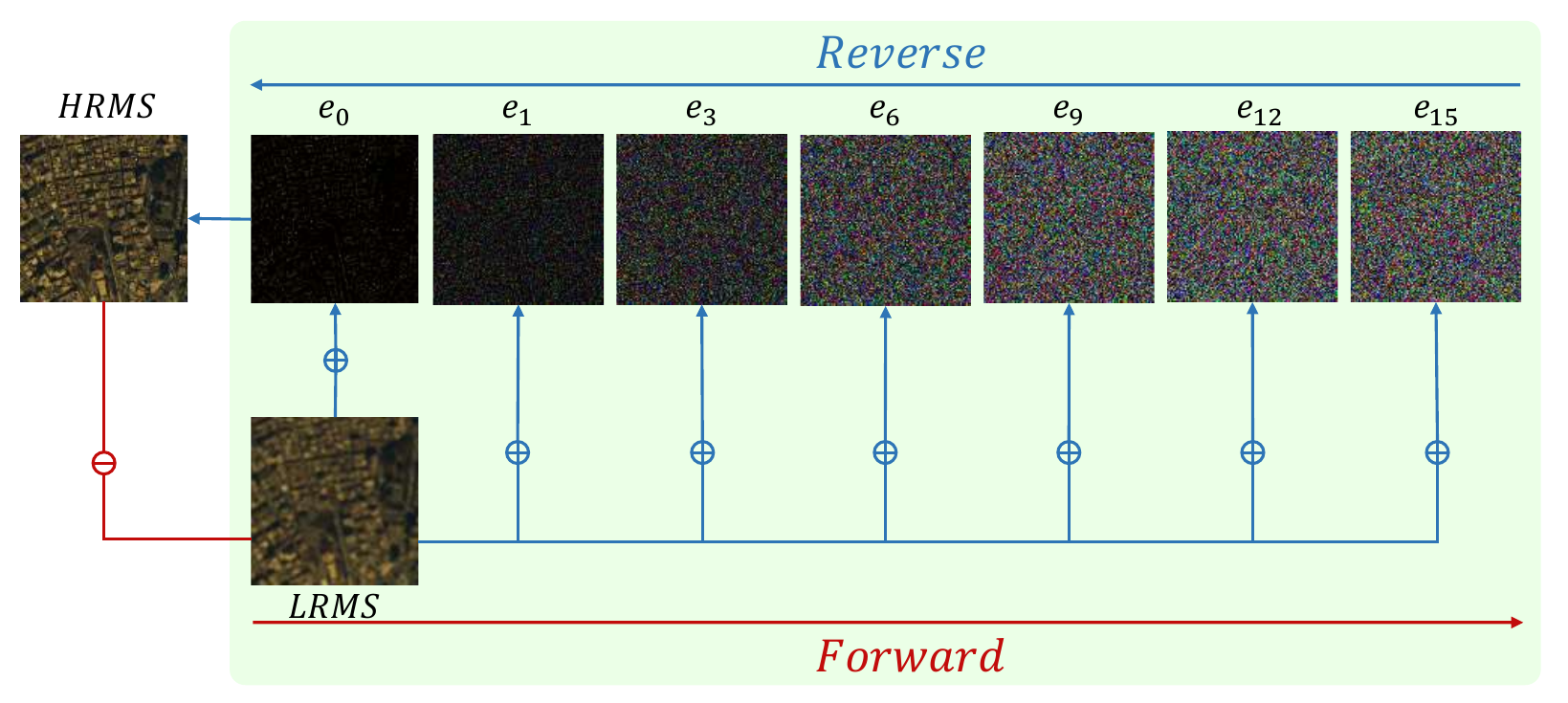}
    \caption{The overview of the proposed diffusion process. It builds up a Markov chain between the $e_T$ and $e_0$}
    \label{fig: diffusion}
\end{figure*}

\begin{figure*}[!ht]
\xdef\xfigwd{\textwidth}
        \centering    
        \includegraphics[width=0.9\linewidth]{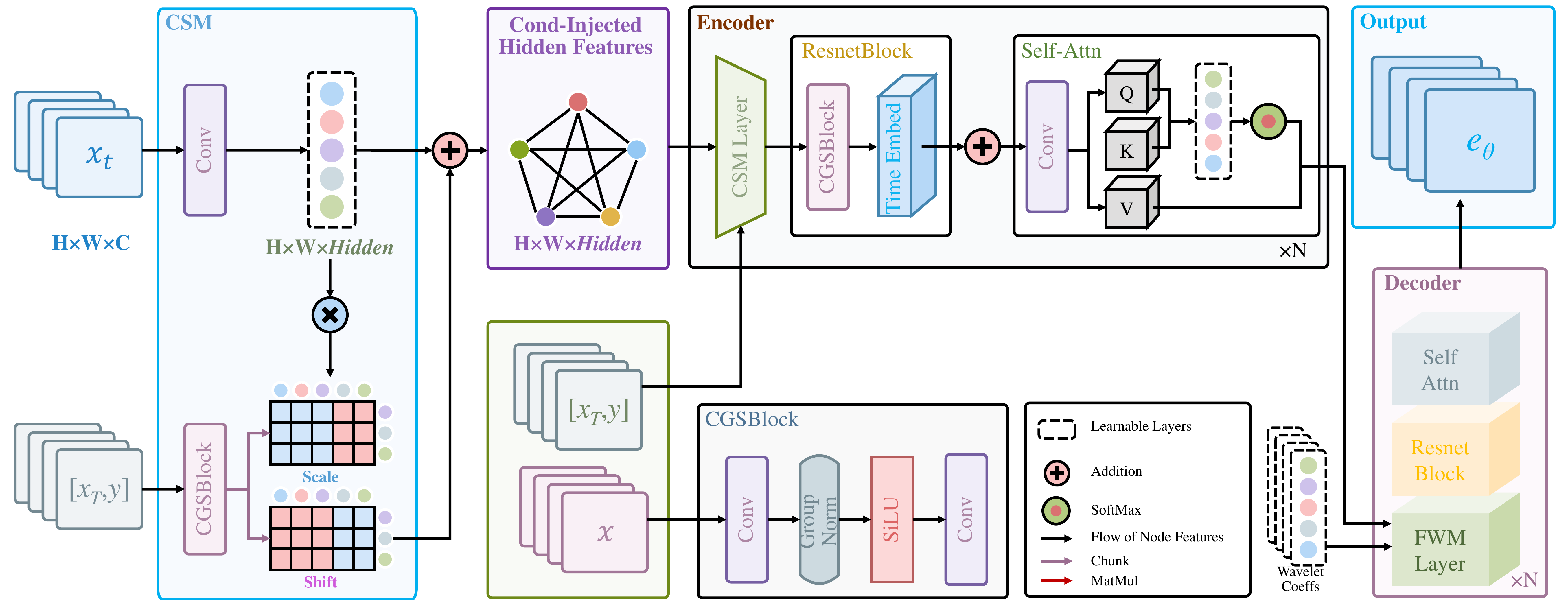}
    \caption{Overview of the architecture of the model. The input gets $x_{t}$ by adding $e_{t}$ to $x_T$. The model employed effective disentangled modulations (i.e., CSM, FWM) to produce the undegraded $e_0$. Finally, $e_0$ is added to LRMS to get the HRMS image.}
    \label{fig: model arch}
\end{figure*}

\setcounter{equation}{0}
\newcommand{\bsy}[1]{\boldsymbol{#1}}
In this section, we first analyze the reason why residual generation makes sense. Then, the design of the diffusion model for pansharpening, whose diffusion process can be divided into forward and reverse processes aiming at generating residual will be introduced. For the latter, we construct the Latent State $x_t$ and the Shallow Cond-Injection by replacing simple convolution with CSM~\cite{cao2024diffusion} in the shallow extraction layer, enabling the model to extract sufficient features. Then, we design the Loss Function, which performs better guidance for the model. Fig.~\ref{fig: diffusion} shows the Markov chain that serves as a bridge between the HRMS and LRMS images through the inference of the residual and Fig.~\ref{fig: model arch} shows the overall architecture of the generation model. We assume that HRMS and LRMS share the same spatial resolution. If necessary, they can be easily achieved by pre-upsampling the low-resolution image $x_T$ using nearest neighbor interpolation. For convenience, we list the related notations in Tab.~\ref{tab: math_notaion}.

\begin{table}[!ht]
    \centering
    \caption{Some Notations Used in This Work.}
    \label{tab: math_notaion}
    % \begin{tabular}{c|@{\hspace{10pt}}l}
    {
    \begin{tabular}{l|l}
    \toprule
        Notation & Explanation \\
        \midrule
        $t$ & The timestep\\
        $e_0$ & The residual between HRMS and LRMS\\
        $x_T$ & The LRMS image\\
        $x_0$ & The HRMS image\\
        $y$ & The PAN image \\
        $DB1(\cdot)$  & The DB1 wavelet decomposition~\cite{db1} \\
        %\epsilon$ & Gaussian noise \\
    \bottomrule
    \end{tabular}}
\end{table}

\subsection{ResPanDiff} \label{sec: ResPanDiff}
\begin{figure}[htb] 
    \centering
    \includegraphics[width=0.5\linewidth]{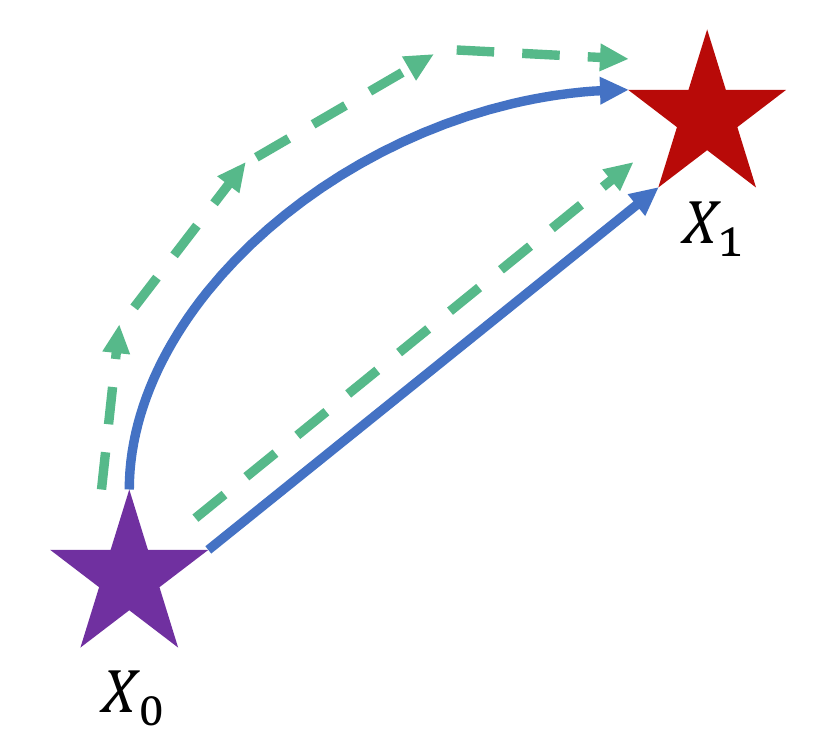}
    \caption{The different lines the transport distribution $X_0$ to $X_1$ follow. The blue one means the original transmission, and the green one represents the simulation line. It is clear that the model takes more times with a smaller step to simulate the curve, while it may perform worse compared to the simulation of the straight line with even only one time and a bigger step.}
    \label{fig: straight line vs.}
\end{figure}

\noindent \textbf{Why does residual generation make sense?}
According to  DPM slover~\cite{lu2022dpm}, the diffusion process can be transformed into SDEs/ODEs equations.
% , which can be specifically expressed as
% %如dpm slover所说，我们知道diffusion过程可以转换成SDEs/ODEs方程，具体可以表示为
% \begin{equation}
%     d \boldsymbol{x}=\left( \boldsymbol{f}_{t} ( \boldsymbol{x} )-{\frac{1} {2}} ( g_{t}^{2}-\sigma_{t}^{2} ) \nabla_{\boldsymbol{x}} \operatorname{l o g} p_{t} ( \boldsymbol{x} ) \right) d t+\sigma_{t} d \boldsymbol{w}_t,
% \end{equation}
% where $d \boldsymbol{w}_t $ is a wiener motion, $g_t$, $\sigma_t$ are determined functions. This equation becomes ODEs when $\sigma_{t} = 0$:
% \begin{equation}
%     d \boldsymbol{x}=\left( \boldsymbol{f}_{t} ( \boldsymbol{x} )-{\frac{1} {2}} ( g_{t}^{2} ) \nabla_{\boldsymbol{x}} \operatorname{l o g} p_{t} ( \boldsymbol{x} ) \right) d t,
% \end{equation}
% %生成目标的扩散过程就可以转换成通过训练神经网络表征drift然后将其用于SDEs/ODEs中以此来模拟过程
By this way, the diffusion process of generating the target can be transformed into representing the drift by training a neural network, which is then used in SDEs/ODEs to simulate the process. In order to speed up this process, Rectified flow~\cite{liu2022flow} shows that the transport distribution $x_T \in \mathbb{R}^{H \times W \times C}$ to $x_0 \in \mathbb{R}^{H \times W \times C}$ should follow straight lines as much as possible, since it will make the simulation more precise at every $t$ step. An obvious example is shown in Fig.~\ref{fig: straight line vs.}, where $X_0 \in \mathbb{R}^{H \times W \times C}$ represents the input image and $X_1 \in \mathbb{R}^{H \times W \times C}$ represents the target image.

To create a straight line, Rectified flow defines a linear interpolation $X_t = tX_1 + (1-t)X_0$ and the mapping corresponds to the following OEDs:
\begin{equation}
    \frac{d}{dt} X_t = v_t(X_t,t).
\end{equation}
The process of transport follows the Euler method:
\begin{equation}
    X_{t+\epsilon}=X_t+v_t(X_t,t)\epsilon ,
\end{equation}
where $\epsilon$ represents a tiny step.

\begin{figure}[!t] 
    \centering
    \includegraphics[width=1.0\linewidth]{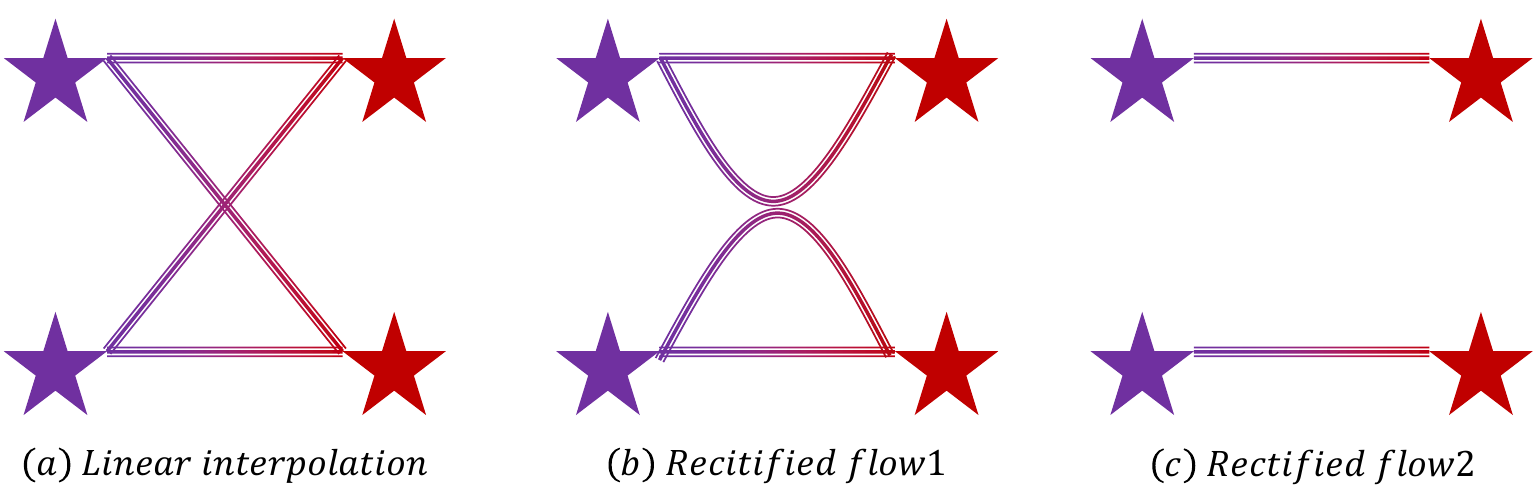}
    \caption{The utility of Rectified flow. We can find that the first Rectified flow makes the trajectory non-crossed, and the second Rectified flow makes it straight.}
    \label{fig: cross}
\end{figure}

However, there are still some problems. When two paths go across, there is no directional guidance at time $t$ along different directions; the solution of the ODE would be non-unique. On the other hand, the paths of the interpolation process $X_t$ may intersect with each other (Fig.~\ref{fig: cross}a), which makes it non-causal.

\begin{figure}
\xdef\xfigwd{\textwidth}
    \centering
        \includegraphics[width=0.45\linewidth]{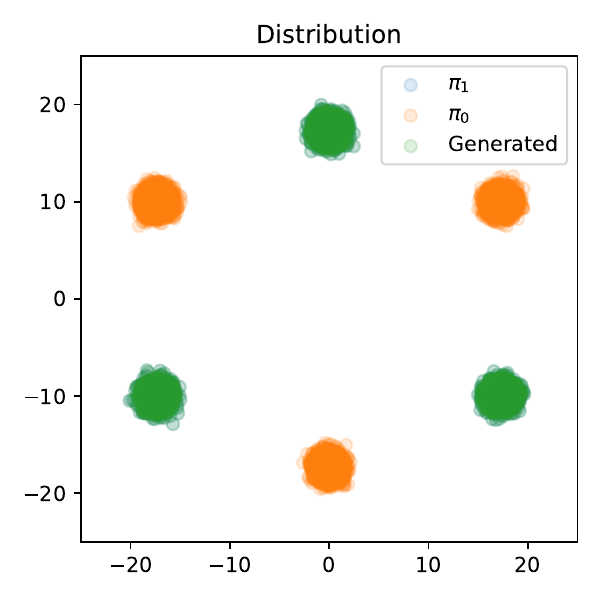}
        \label{1a}\hfill    
        \includegraphics[width=0.45\linewidth]{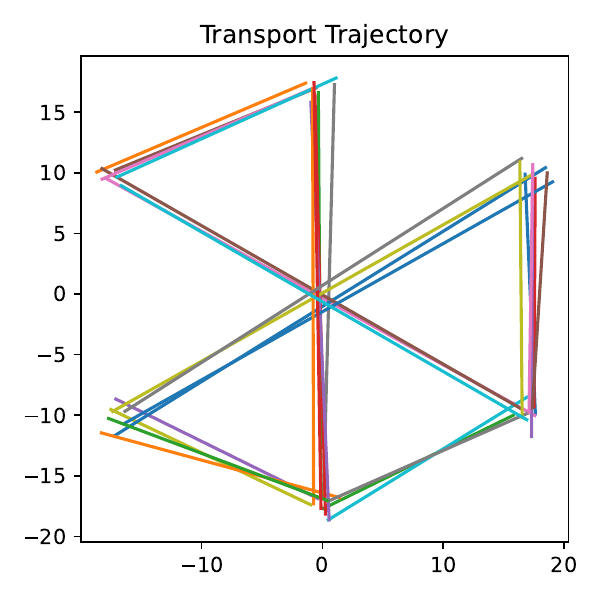}
        \label{1b}\
    \centering
        \includegraphics[width=0.45\linewidth]{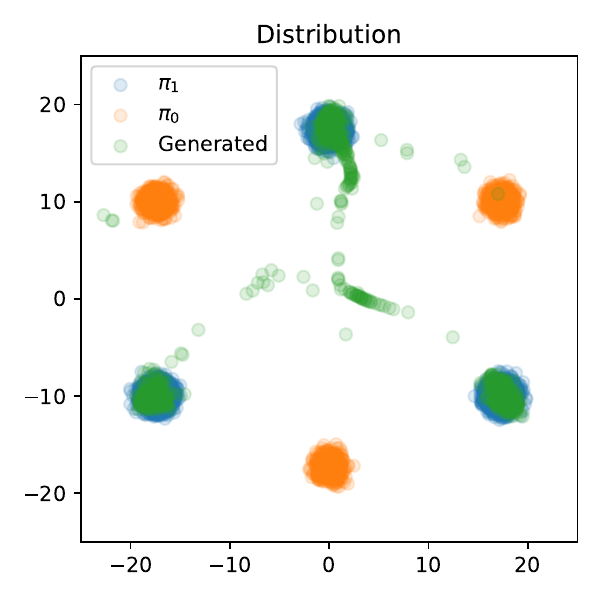}
        \label{1c}\hfill
        \includegraphics[width=0.45\linewidth]{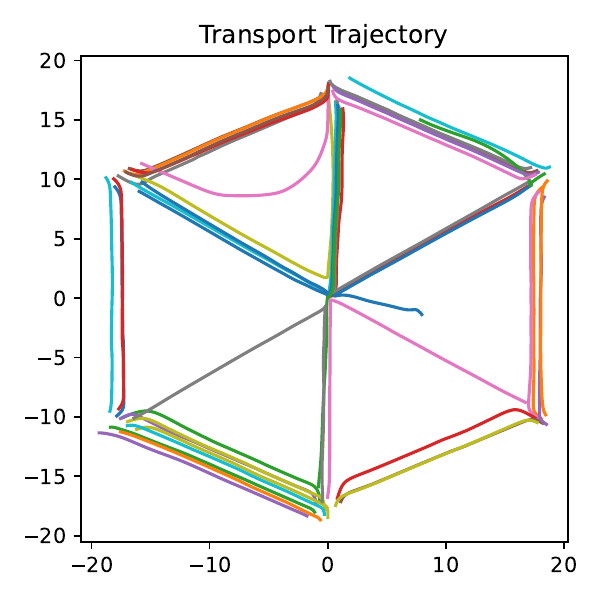}
        \label{1d}\
    \centering
        \includegraphics[width=0.45\linewidth]{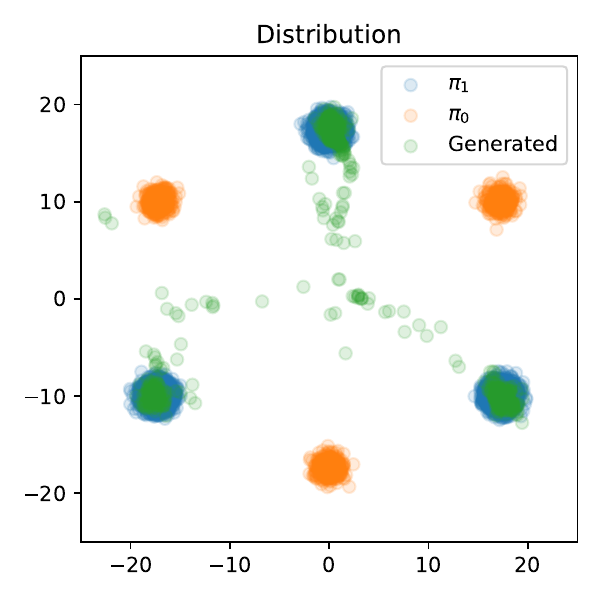}
        \label{1e}\hfill
        \includegraphics[width=0.45\linewidth]{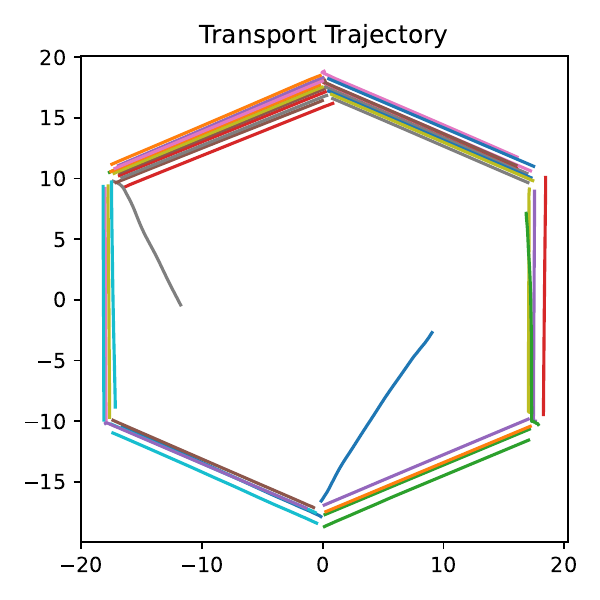}
        \label{1f}
    \caption{The different stages of the Rectified flow. The first column represents the data distribution of $X_0 \sim \pi_0$, $X_1 \sim \pi_1$ and the generated $X$ through the generation path. The second column represents the transport trajectory from $X_0$ to $X_1$ at each stage.}
    \label{fig: flow stage}
\end{figure}

Through mathematical proof in Rectified flow, the following approach is proposed to avoid path intersections and straighten the trajectories as much as possible, as detailed below:
\begin{equation}
    \begin{array} {l} {{\operatorname* {m i n}_{v} \int_{0}^{1} \mathbb{E}_{X_{0}, X_{1}} \left[ | | ( X_{1}-X_{0} )-v ( X_{t}, t ) | |^{2} \right] dt,}}  \\ \end{array} 
    \label{eq: flow_1}
\end{equation}
where $X_{t}=t X_{1}+( 1-t ) X_{0}$ and ${X_{0} \sim\pi_{0}, X_{1} \sim\pi_{1}}$. After getting 1-Rectified flow by the optimization, the model can make lines non-crossive while ensuring
that they trace out the same marginal distributions, as is shown in Fig~\ref{fig: cross}b. They then train the 2-Rectified flow by following the optimization:
\begin{equation}
    \begin{array} {l} {{\operatorname*{min}_{v} \int_{0}^{1} \mathbb{E}_{X_{0}, X_{1}} \left[ | | ( X_{1}-X_{0} )-v ( X_{t}, t ) | |^{2} \right] d t, }} \\ \end{array} 
\end{equation}
where $X_{t}=t X_{1}+( 1-t ) X_{0}$, ${X_{0} \sim\pi_{0}, X_{1} \sim flow1(X_0)}$ and $flow1$ represents the trained 1-Rectified flow.
After the two stages, the path has become straight (Fig.~\ref{fig: cross}c). Specifically, what the two stages have to do is shown in Fig.~\ref{fig: flow stage}
% \begin{figure}[!t] 
%     \centering
%     \includegraphics[width=1.0\linewidth]{figs/tmp/tmp_flow.pdf}
%     \caption{The different stage of the Rectified flow. (a, (c, (e represent the data distribution of $X_0 \sim \pi_0$, $X_1 \sim \pi_1$ and the generated $X$ through the generation path. (b, (d, (f represent the transport trajectory from $X_0$ to $X_1$ at each stage.}
%     \label{fig: flow stage}
% \end{figure}

% \begin{figure}[htbp]
% \centering
% \subfigure[pic1.]{
% \includegraphics[width=5.5cm]{figs/method/Dist_Rf1.pdf}
% }
% \quad
% \subfigure[pic2.]{
% \includegraphics[width=5.5cm]{figs/method/Trans_Rf1.pdf}
% }
% \quad
% \subfigure[pic3.]{
% \includegraphics[width=5.5cm]{figs/method/Dist_Rf2.pdf}
% }
% \quad
% \subfigure[pic4.]{
% \includegraphics[width=5.5cm]{figs/method/Trans_Rf2.pdf}
% }
% \caption{ pics}
% \end{figure}

% \begin{figure}[t]
% \centering  %图片全局居中
% \subfigure[]{
% \label{Fig.sub.1}
% \includegraphics[width=0.49\linewidth]{figs/exp/Dist_wv3.pdf}}
% \subfigure[]{
% \label{Fig.sub.2}
% \includegraphics[width=0.49\linewidth]{figs/exp/Trans_wv3.pdf}}
% \\
% \subfigure[]{
% \label{Fig.sub.1}
% \includegraphics[width=0.49\linewidth]{figs/exp/Dist_wv3.pdf}}
% \subfigure[]{
% \label{Fig.sub.2}
% \includegraphics[width=0.49\linewidth]{figs/exp/Trans_wv3.pdf}}
% \caption{\ref{Fig.sub.1} represents the data distribution of $LRMS \sim \pi_0$, $HRMS \sim \pi_1$ and the generated $SR$ through the generation path, \ref{Fig.sub.2} represents the transport trajectory from $LRMS$ to $HRMS$.}
% \label{1}
% \end{figure}

As expressed in Rectified flow, 1-Rectified flow ensures that paths no longer cross, and 2-Rectified flow makes paths straight. In this way, the speed of generation has been significantly improved. \textit{Returning to the MSIF task, can we refine this approach to not only effectively improve sampling speed but also make it more elegant and streamlined compared to previous methods?}

\begin{figure}[t]
\centering  %图片全局居中
{
\label{Fig.sub.1}
\includegraphics[width=0.49\linewidth]{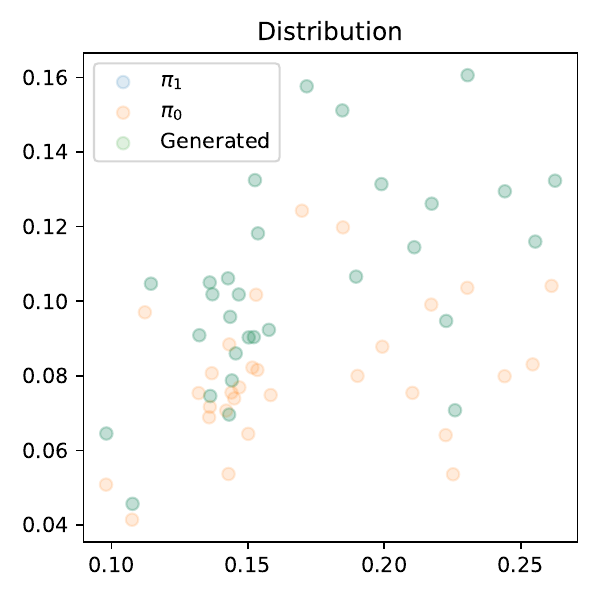}}{
\label{Fig.sub.2}
\includegraphics[width=0.49\linewidth]{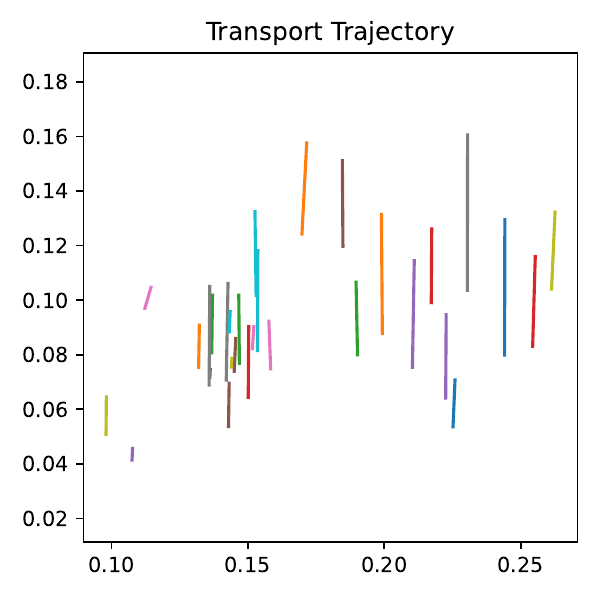}}
\caption{The distribution of the WV3 dataset and the transport trajectory from LRMS to HRMS. The first column represents the data distribution of $LRMS \sim \pi_0$, $HRMS \sim \pi_1$ and the generated $SR$ through the generation path. The second column represents the transport trajectory from $LRMS$ to $HRMS$}
\label{fig: wv3 transport}
\end{figure}

Compared to image generation tasks, in the MSIF task, $X_0$ and $X_1$ are no longer randomly paired but instead have a fixed relationship. Specifically, $X_1$  corresponds to HRMS and $X_0$ corresponds to LRMS, which is a degraded version of $X_1$. \textit{Given this, could we hypothesize that the crossovers generated along the path from LRMS to HRMS are fewer?} Based on this assumption, we plot the distribution of WV3 data and the generation path from LRMS to HRMS, and our hypothesis has been confirmed, which is shown in Fig.~\ref{fig: wv3 transport}.
In Fig.~\ref{fig: wv3 transport}, we can find that the transport trajectory from $LRMS$ to $HRMS$ is straight with minimal cross-interference. This suggests that in the MSIF task, we do not need to go through stages like the 1-Rectified flow to avoid cross-interference; instead, we can find a straightforward generation path. So, how do we determine this path? We know that the path is defined by the model-generated vector $v(X_t,t)$ during training, which means that once we determine $v(X_t,t)$, we can obtain the desired path.

We observe that in Eq.~\eqref{eq: flow_1}, $X_1 - X_0$  precisely corresponds to the definition of the residual $e_0 = X_1 - X_0$. Therefore, by replacing $X_1 - X_0$ with the residual $e_0$ as the optimization target, we can obtain the desired $v(X_t,t)$, which in reality is the residual $e_0$ that we need to fit. So the ODEs become:
\begin{equation}
    \frac{d}{dt} X_t = \hat{e}_0(X_t,t),
\end{equation}
where $\hat{e}_0(X_t,t)$ represents the predicted residual by the model, which means that the generation target of the model becomes $\hat{e}_0(X_t,t)$ rather than $v(X_t,t)$. In this way, the optimal target becomes:
\begin{equation}
     \begin{array} {l} {{\operatorname* {min}_{v} \int_{0}^{1} \mathbb{E}_{X_{0} \sim\pi_{0}, X_{1} \sim\pi_{1}} \left[ | | e_{0} - \hat{e}_0 ( X_{t}, t ) | |^{2} \right] d t.}}  \\ \end{array} 
\end{equation}
From this, we can conclude that residual training not only significantly enhances sampling speed but also ensures accuracy. We find that no one have built a Markov chain aiming at residual. This concept inspires us to devise a Markov chain for the transmission of residuals. Specifically, the Markov chain orchestrates the learned residual transitions represented by $q ({e}_{0}|{x}_{T}, y) :=\int q( {e}_{0 : T}|{x}_{T}, y) \, d {e}_{1 : T} $ staring at $e_T$, where $e_1, \cdots, e_T $ are latent variables sharing the same dimensionality as the data $ e_0 \sim q(e_0)$.

\noindent \textbf{{Forward Process}}
To effectively generate residual, we define $x_0 = X_1$ and $x_T = X_0$, the ODEs becomes:
\begin{equation}
       -\frac{d}{dt} x_t = e_0.
\end{equation}
By considering each step $\epsilon$ as $t$-th step, the process of transport becomes
\begin{equation}
    x_{t}=x_{t-1}-\alpha_t e_0,
\end{equation}
where $\alpha_t$ represents the step size at $t$-th step. By simultaneously subtracting $x_T$ from both sides, we get the formulation for $e_t$:
\begin{equation}
    e_{t}=e_{t-1}-\alpha_t e_0.
\end{equation}
Having this, we propose a Markov chain from $e_T$ to $e_0$, which is the forward process of ResPanDiff. In the forward process, the foundational concept involves transforming $e_0$ into $e_T$ by employing a Markov chain of length $T$. Initially, it is essential to establish a variance schedule $\alpha_{1},\cdots,\alpha_{T}$ where $\alpha_{1}\rightarrow 0$ and $\alpha_{T}\rightarrow 1$. Subsequent to this setup, the transition distribution we propose is defined as follows:
\begin{equation} \label{eq: q_one_sample}
q(e_{t}|e_{t-1}, e_{0}) \backsim {\cal N}(e_{t-1}-\alpha_{t}e_{0}, \kappa^2\alpha_{t}),~t=1, 2, \cdots, T,
\end{equation}
where $\kappa$ is a hyper-parameter which controls the noise variance. Notably, we demonstrate that the marginal distribution at any timestep $t$ is analytically integrable, expressed as:
\begin{equation} \label{eq: one_forward_step}
q(e_{t}|e_{0}) \backsim {\cal N}((1-\overline{\alpha}_{t})e_{0},\kappa^2\overline{\alpha}_{t}),~t=1,2,\cdots,T,
\end{equation}
where $\overline{\alpha}_{t}$ represents for $\displaystyle\sum_{i=1}^{t} \alpha_{i}$. With this marginal distribution, we are capable of determining $e_t$ at any timestep $t$ in a single forward step.

\noindent \textbf{Reverse Process}
The reverse process focuses on estimating the posterior distribution $p_{\theta}(e_{0}|x_{T},y)$. This estimation is derived through the following integral:
\begin{equation}
    p_{\theta}(e_{0}|x_{T},c) = \int p(e_{T} | x_{T}) \prod_{t=1}^{T} p_{\theta}(e_{t-1}|e_{t},x_{T},c) \, \mathrm{d} {e}_{1:T},
    \label{eq: p_e0|xT}
\end{equation}
where $p(e_{T}|x_{T})$ approaches zero. $p_{\theta}(e_{t-1}|e_{t},x_{T},c)$ represents the transition of latent variables from $e_{t}$ to $e_{t-1}$, conditioned on inputs $x_{T}$ and $c$ at each timestep $t$. To be more specific, $c$ is obtained by DB1 wavelet decomposition~\cite{db1} to decompose both $x_T$ and $y$ into four components, respectively, which include one main component and three detail components in the horizontal, vertical and diagonal directions.
The process can be expressed as follows:
\begin{equation}
\begin{aligned}
    \bsy{LL}_{x_T}, \bsy{LH}_{x_T}, \bsy{HL}_{x_T}, \bsy{HH}_{x_T}
    &=\text{DB1}(x_T), \\
    \bsy{LL}_{y}, \bsy{LH}_{y}, \bsy{HL}_{y}, \bsy{HH}_{y}
    &=\text{DB1}(y),
\end{aligned}
\end{equation}
where $\bsy{LL}\in \mathbb R^{\frac H 2 \times \frac W 2 \times d}$ denotes the low-frequency main component, $\bsy{LH}\in \mathbb R^{\frac H 2\times \frac W 2 \times d}$, $\bsy{HL}\in \mathbb R^{\frac H 2\times \frac W 2 \times d}$, and $\bsy{HH}\in \mathbb R^{\frac H 2\times \frac W 2 \times d}$ denote the detail components in the horizontal, vertical, and diagonal directions, respectively. $d$ is the dimension. Then the attention operation can be used in FWM for modulation.

In particular, the tractability of $p_{\theta}(e_{t-1}|e_{t},x_{T},c)$, conditioned on $x_{T}$ and $c$, is given by:
\begin{equation}
    p_{\theta}(e_{t-1}|e_{t},x_{T},c) = \mathcal{N}(e_{t-1} ; \mu_{\theta}(e_{t}, x_{T}, c, t), \Sigma_{\theta}(e_{t}, x_{T}, t)),
\end{equation}
where $\theta$ denotes the learnable parameters of the model $\mu_{\theta}(e_{t}, x_{T}, c, t)$ and $\Sigma_{\theta}(e_{t}, x_T, t)$ are the mean and the covariance, respectively. To optimize $\theta$, we aim to minimize the negative evidence lower bound:
\begin{equation} \label{eq:KL}
    \operatorname*{min}_{\theta} \sum_{t} D_{\mathrm{KL}} \left[ q(e_{t-1}| e_{t}, e_{0}) \Vert p_{\theta}(e_{t-1} | e_{t}, x_{T}, c) \right],
\end{equation}
where $D_{\mathrm{KL}}[\cdot \Vert \cdot]$ denotes the Kullback-Leibler divergence, which quantifies the deviation of $p_{\theta}(e_{t-1} | e_{t}, x_{T}, c)$ from the forward process posteriors $q(e_{t-1}|e_{t},e_{0})$.

According to Bayes' theorem, the target distribution $q(e_{t-1}|e_{t},e_{0})$ in Eq.~\eqref{eq:KL} is formulated as follows:
\begin{equation} \label{eq:Bayes}
    q(e_{t-1}|e_{t},e_{0}) \propto q(e_{t}|e_{t-1},e_{0}) q(e_{t-1}|e_{0}).
\end{equation}

Upon further substitution of Eq.~\eqref{eq: q_one_sample} and Eq.~\eqref{eq: one_forward_step} into Eq.~\eqref{eq:Bayes}, the expression for $q(e_{t-1}|e_{t},e_{0})$ becomes tractable and can be explicitly represented as:
\begin{equation} \label{eq:reverse}
    p(e_{t-1}|e_{t},e_{0}) \sim \mathcal{N}\left(\frac{\overline{\alpha}_{t-1}}{\overline{\alpha}_t} e_{t} + \frac{\alpha_t}{\overline{\alpha}_t} e_{0}, \kappa^2 \frac{\overline{\alpha}_{t-1} }{\overline{\alpha}_t}\alpha_t\right).
\end{equation}
A detailed derivation of this formulation is presented in \ref{appendix: math}. This establishes the training target, but an issue remains. As previously noted, $p(e_{T}|x_{T}) \rightarrow 0$ indicates that $p(e_{T}|x_{T})$ provides little to useful information for model learning, leading to inefficiencies in early-stage model inference. To address this, we introduce a Latent State $x_t = x_T + e_t$ and utilize $x_t$ instead of $e_t$ as the model input. Consequently, the mean parameter $\mu_{\theta}(x_{t}, t, c)$ and variance $\Sigma_{\theta} ({e}_{t}, x_T, t)$ are reparameterized as follows:
\begin{equation} \label{eq:miu}
\mu_{\theta}(x_{t}, c, t)=\frac{\overline{\alpha}_{t-1}}{\overline{\alpha}_{t}}e_{t}+\frac{\alpha_{t}}{\overline{\alpha}_{t}}f_{\theta}(x_{t}, c, t),
\end{equation}
\begin{equation}
    \Sigma_{\theta} (x_t, t) = \kappa^2 \frac{\overline{\alpha}_{t-1} }{\overline{\alpha}_t}\alpha_t.
\end{equation}

Based on Eq.~\eqref{eq:miu}, The optimization for $\theta$ is achieved by minimizing the negative evidence lower bound in Eq.~\eqref{eq:KL}, namely,
\begin{equation} \label{eq: loss_pix}
    \operatorname* {min}_{{\theta}} \sum_{t} \| f_{{\theta}} ( {x}_{t}, c, t )-{e}_{0} \|_{2}^{2}.
\end{equation}

In pursuit of this objective, the target residual $e_0$ is sampled in discrete steps $T$ throughout the model training process. This methodical sampling is critical for effectively training the model to approximate the target output as accurately as possible. Sequential sampling at each of these $T$ steps ensures that the model incrementally adjusts its parameters to better predict the target residual, thus enhancing overall training efficacy.

It is important to note that none of the existing diffusion models are capable of achieving this. Many diffusion models have introduced the concept of residuals and applied it during the diffusion process~\cite{yue2024resshift, liu2024residual}. However, none of the diffusion models focus on the inference of the residuals themselves; in their work, residuals are merely an intermediary quantity in the diffusion process rather than the target of inference. Ultimately, their inference process results in a complete image $x_0$. Furthermore, aside from differing definitions of the residual($e_0 = LRMS - HRMS$ in \cite{yue2024resshift, liu2024residual} if used in pansharpening task while $e_0 = HRMS - LRMS$ in ours), the model in ResShift~\cite{yue2024resshift} predicts $\hat{x}_0$, while the model in ResDiff~\cite{liu2024residual} predicts both the residual~$e_\theta$ and noise~$\epsilon_\theta$, whereas the model focuses solely on predicting the residual~$\hat{e}_0$ itself.

\subsection{Latent State: $x_t$} \label{sec: input:xt} %想个名字 
% In the previous work \cite{cao2024diffusion}, DDIF directly uses the residual image instead of the low-resolution image as input during the training phase, while in the sampling phase, it starts from Gaussian noise, which leads to the issue that the model gets less information. 
% In DDIF, FWM which supplements frequency domain information in the decoder layer through wavelet transformation and attention mechanisms to generate high-quality residuals. Although this method addresses the problem of insufficient information when using residuals as input, the introduction of additional information and attention mechanisms has resulted in significant memory usage and decreased speed during the generation process.

%作为输出，e_t具有很好的性质，然而作为输入的话，信息过于稀疏不利于提取，不如x_t
As an output, $e_{t}$ possesses favorable properties. The model needs to generate less information, which enables it to focus more on the generation of residual information. However, when $e_t$ is employed as an input, its sparseness hinders efficient information extraction, unlike utilizing the full image. While a full image as input seems logical, it presents challenges within the diffusion model, as diffusion processes involve images at different time steps $t$. Based on this consideration, we design a novel input that allows the model to receive sufficient information at the input stage and focus on generating residual information for the diffusion model. We define:
\begin{equation}
    x_{t} = e_{t} + x_T.
\end{equation}
With $x_t$ as the input, it effectively addresses the issues mentioned above, enabling the model to not only acquire sufficient information at the input stage but also exhibit desirable properties during the diffusion process.

%从频域和像素两个角度说明
\subsection{Shallow Cond-Injection (SC-I)} \label{sec: generate model}
% When processing images from their own dimensions to hidden dimensions, most models employ a single shallow feature extraction to extract the shallow feature to the hidden dimensions, followed by in-depth processing through the model. However, in MSIF task, it may not be suitable since there's another image pan which offering the spatial information. Hence, we utilize a shallow cond-injection to replace the original shallow extraction layer. Specifically, we generate the corresponding style feature by CSM proposed in DDIF rather than just a deep feature by a simple convolution, which has significantly improved the performance. 
In the process of transforming images from their native dimensions to hidden dimensions, most models typically employ a simple shallow feature extraction step to transition these features into hidden dimensions. This is then followed by more comprehensive processing within the model. However, in the MSIF task, this approach may not be suitable due to the presence of another image PAN, which provides spatial information. To address this, we employ a shallow conditional injection method in place of the original shallow extraction layer. Specifically, instead of relying solely on deep feature extraction through a simple convolution, we generate the Cond-Injected Hidden Feature $\in \mathbb{R}^{H \times W \times Hidden}$ using CSM as proposed in DDIF.  This strategy has substantially enhanced performance.

\subsection{Loss Function}
The loss function plays a critical role in guiding model training. Even if a model has an excellent structure, it is difficult to fully realize its superior performance without a well-designed loss function. For residual generation tasks, due to their unique characteristics, we make specific adjustments to the loss function, tailored to the nature of residuals. Compared to conventional loss functions designed for image generation, these adjustments enable the model to perform significantly better when generating residuals.
\subsubsection{Residual Loss}
\begin{figure*}[!ht]
    \centering  %图片全局居中
    {
    \includegraphics[width=0.49\linewidth]{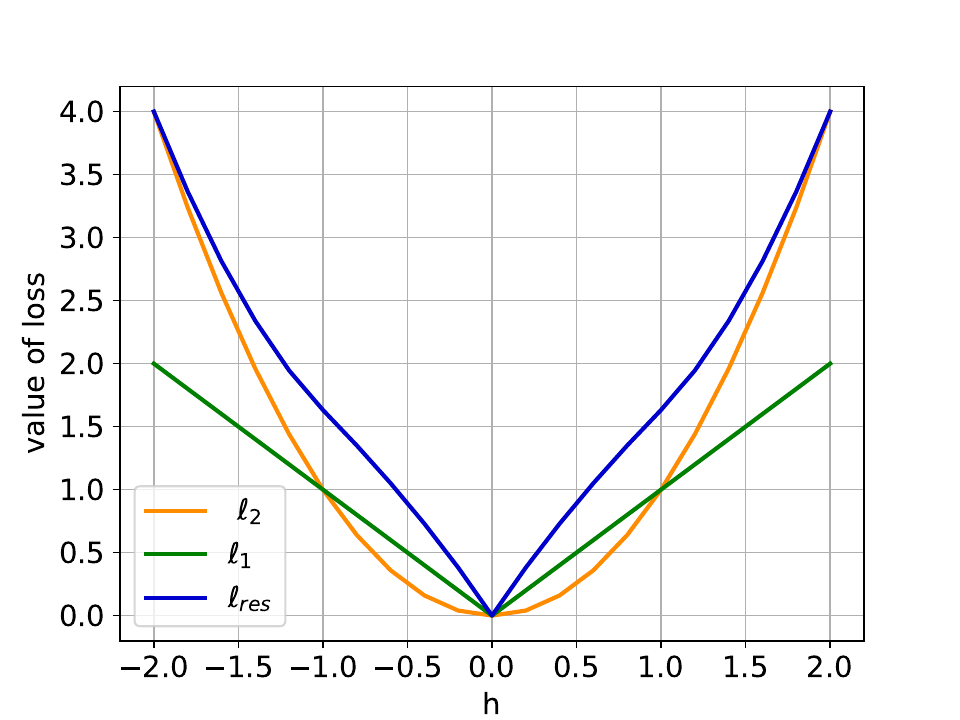}}\hspace{-5mm}{
    \includegraphics[width=0.49\linewidth]{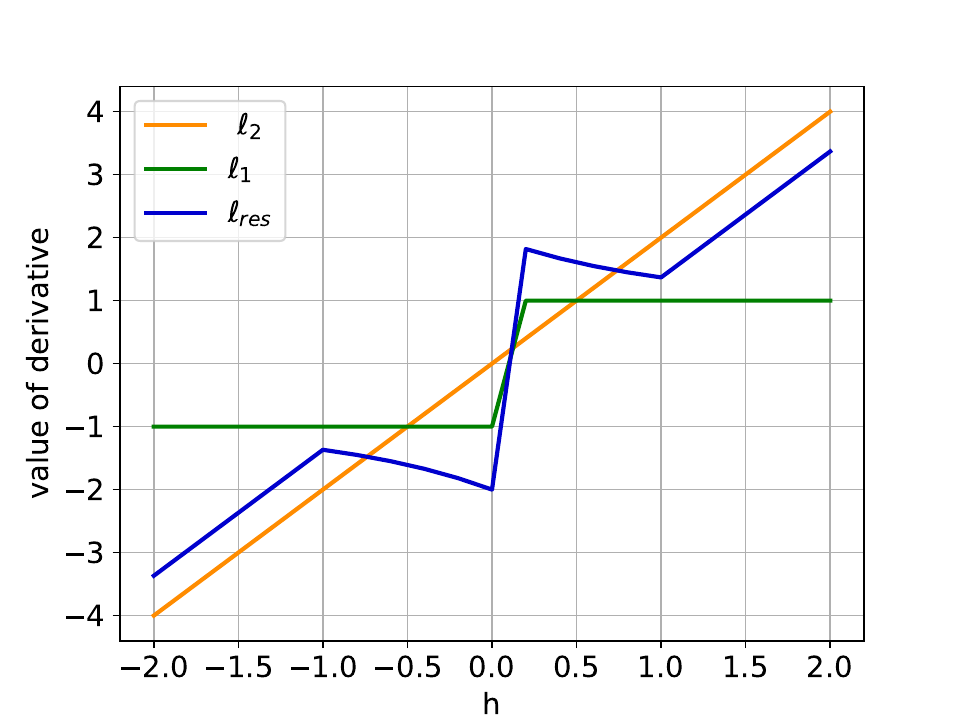}}
    \caption{The left one represents the figs of different losses. Comparing with $\ell_1$ and $\ell_2$, $\ell_{res}$ is more precipitous in the domain of zeroes, which enables better guidance of the model for the residual generation task. The right one represents the derivatives of different losses. Comparing with $\ell_1$ and $\ell_2$, the derivatives of $\ell_{res}$ in the domain of zeros are neither too small nor constant, which enables better guidance of the model for the residual generation task.}
    \label{fig: losses and derivatives}
\end{figure*}

The $\ell_1$ loss provides a stable gradient regardless of the input values, which often enhances the robustness of the model. However, since changes in the input have no impact on the gradient, it lacks sensitivity in tasks like residual generation, where the differences are small. On the other hand, the $\ell_2$ loss function is sensitive to outliers and offers a relatively stable solution, but in residual generation tasks where inputs are typically concentrated around zero, the gradients produced by $\ell_2$ loss are always very small, limiting its effectiveness as a guiding force. To solve these issues, we design a new loss as follows:
\begin{equation}
\mathcal{L}_{res} = \left\{\begin{matrix}
\begin{aligned}
 &|h|+(1-e^{-|h|})&&|h|<1,
\\&(|h|+a)^2+b &&|h|>=1,
\end{aligned}
\end{matrix}\right.
\end{equation}
where $h$ represents $(e_0 - f_{{\theta}} ( {e}_{t}, {x}_{T}, c, t ))$, $a=1/2e-1/2$, $b=7/4-3/2e-1/4e^2$. The comparison among $\ell_1$, $\ell_2$ and $\ell_{res}$ is shown in Fig.~\ref{fig: losses and derivatives}.

\subsubsection{Boundary Penalty}
In self-supervised training, the model focuses on the predicted residuals rather than the original image. This introduces a key distinction: while the original image typically has a fixed range of [0, 1], the residuals often exhibit a more dynamic range. Specifically, the maximum value of the residuals is less than 1, and the minimum value can be negative. However, when we obtain the final output image $sr$ by adding the residuals to the LRMS, the resulting image is expected to remain within the range [0, 1]. This motivates us to design a boundary penalty that constrains the range of the predicted residual $\hat{e}_0$. The boundary penalty can be formulated as follows:
\begin{equation}
   \mathcal{L}_p = Mean(\mathcal{F}_c(\hat{e}_0-max(e_0))+\mathcal{F}_c(min(e_0)-\hat{e}_0)),
\end{equation}
where $Mean( \cdot )$ denotes the average pooling operation. $\mathcal{F}_c$ represents the clamp operation to ensure the loss is non-negative.
% $min(\cdot)$ denotes finding the minimum value, and $max(\cdot)$ denotes finding the maximum value.

To sum up, the final loss function can be expressed as follows:
\begin{equation}
    \mathcal{L}_{full} = \mathcal{L}_{res} + \gamma\mathcal{L}_p,
\end{equation}
where $\gamma$ is a penalty parameter, which is set to be 10000. In practice, we empirically find that the penalty parameter $\gamma$ results in an evident improvement in performance. With these two losses, the model can be guided well to tackle the residual generation task.
The overall training and sampling stages are presented in Alg.~\ref{alg. train} and \ref{alg. sample}.

\begin{algorithm}[!ht]
    \renewcommand{\algorithmicrequire}{\textbf{Input:}}
	\renewcommand{\algorithmicensure}{\textbf{Output:}}
	\caption{Training stage of ResPanDiff}
    \label{alg. train}
    \begin{algorithmic}[1] % 控制是否有序号
        \REQUIRE  LRMS image $ x_{T}$, PAN image $ y$, HRMS image $ x_0$, Diffusion model $ f_\theta$ with its parameters $\theta$, timestep $t$; % input 的内容
	    \ENSURE Optimized diffusion model $\hat{e}_0$; % output 的内容
        
        \STATE $c\leftarrow  y,  x_{T}, \text{DB1}( y,  x_{T})$;
        \STATE $c = a+b;$
        
        \WHILE {until convergence}
            \STATE $t\leftarrow\ $Uniform($0, T$);
            \STATE $e_{0} = x_{0} - x_{T}$;
            \STATE $e_{t} \leftarrow (1-\overline{\alpha}_{t})e_{0}+\kappa {\overline{\alpha}_{t}}$;
            \STATE $x_t\leftarrow e_{t}+x_{T}$;
            \STATE $\hat{ e}_0\leftarrow {f_\theta}( x_t,  c)$;
            \STATE $\theta \leftarrow \nabla_\theta \ell_{full}(\hat{ e}_0,  e_0)$.
        \ENDWHILE
    \end{algorithmic}
\end{algorithm}

\begin{algorithm}[!ht]
    \renewcommand{\algorithmicrequire}{\textbf{Input:}}
	\renewcommand{\algorithmicensure}{\textbf{Output:}}
	\caption{Inference stage of ResPanDiff}
    \label{alg. sample}
    \begin{algorithmic}[1] % 控制是否有序号
        \REQUIRE  LRMS image $ x_{T}$, PAN image $ y$, trained diffusion model $ f_{\theta^*}$ with its parameters $\theta^*$, sampled residual $ e_t$ at timestep $t$; % input 的内容
	    \ENSURE Sampled residual $e_0$ and the target image $ x_0$; % output 的内容
        
        \STATE $t\leftarrow T$; 
        \STATE $c\leftarrow  y,  x_{T}, \text{DB1}( y,  x_{T})$;
        
        \WHILE {$t\geq 0$}
            \STATE $\hat{e}_{0} \gets f_{\theta^{*}}(x_{t},c)$;
            \STATE $e_{t-1} \backsim p(e_{t-1}|e_{t},\hat{e}_{0})$;
            \STATE $x_{t} \gets e_{t} + x_{T}$;
            \STATE $t\gets t-1$.
        \ENDWHILE

        \STATE $x_{0} \gets e_{0}+x_{T}$.
    \end{algorithmic}
\end{algorithm}

%         \\ 
%         \\
%         \While{}
%         {
%         $\hat{e}_{0} \gets f_{\theta^{*}}(x_{t},c)$; \\
%             $e_{t-1} \backsim p(e_{t-1}|e_{t},\hat{e}_{0})$; \rtcp{Eq.~\eqref{eq:reverse}} \\
%             $x_{t} \gets e_{t} + x_{T}$;\\
%             $t\gets t-1$.
%         }
%         $x_{0} \gets e_{0}+x_{T}$.
%     \end{algorithm}\\

%% file: sections/exps.tex
\section{Experiments}
\label{sec:experiments}
% \subsection{Experimental Settings}
In this section, we will provide a comprehensive overview of the implementation details and compare them with several specialized models for some classic datasets, aiming to demonstrate their effective characteristics for this work. 

\subsection{Noise Schedule}
\begin{figure}[!ht]
    \includegraphics[width=\linewidth]{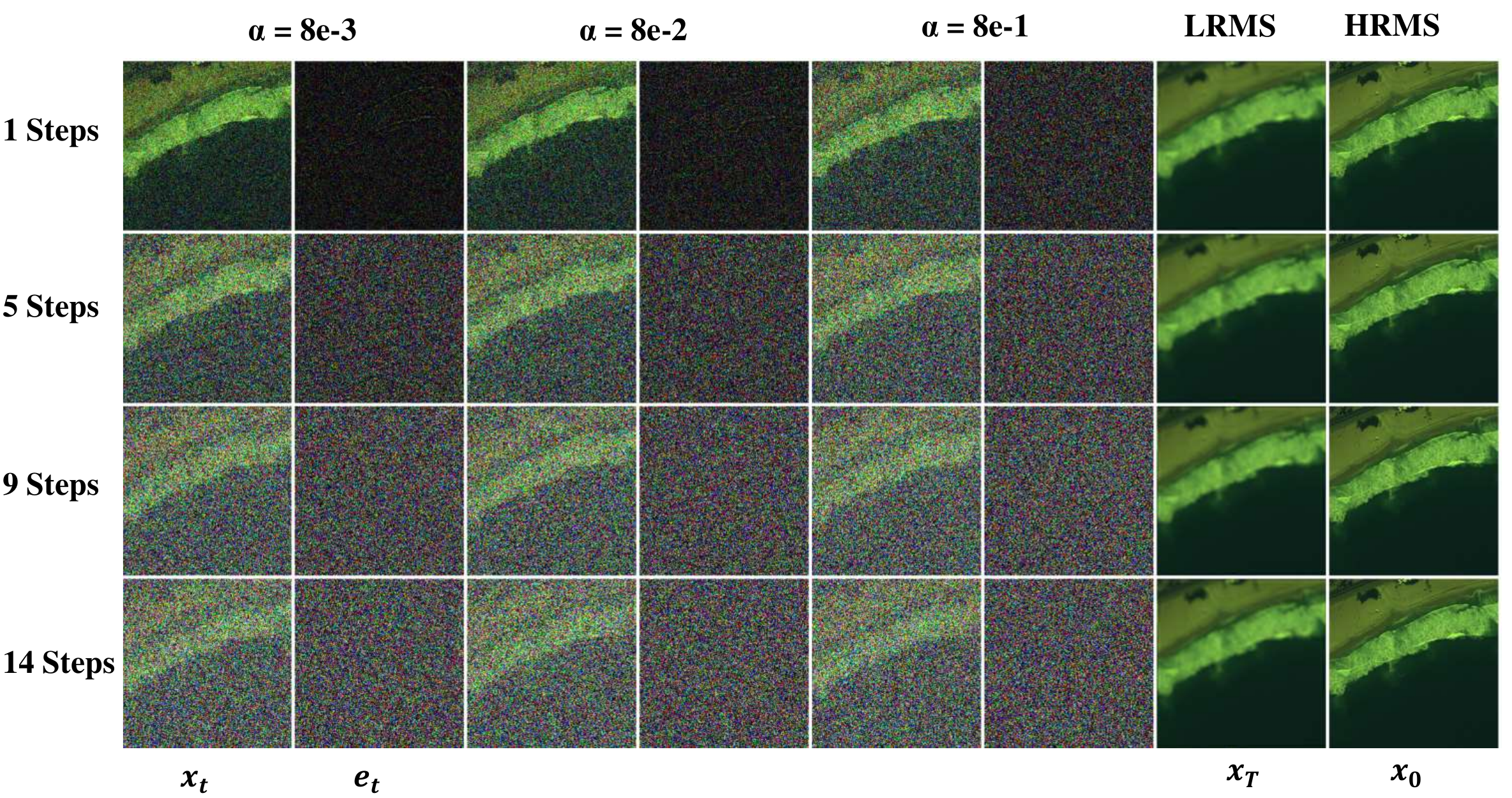}
    \caption{Illustration of the noise schedule for ResPanDiff. We dynamically adjust the parameter of $\alpha$ from $8e-3$ to $8e-1$. These images are obtained in timesteps of 1, 5, 9 and 14, demonstrating the effect of different values of $p$ while keeping $\kappa = 1$ and $T = 15$ fixed.}
    \label{fig: schedule_cosine}
\end{figure}

For the diffusion model, we choose a cosine schedule~\cite{improved_ddpm} for $\alpha_t$, see as follows,
\begin{equation}
    \bar \alpha_t=1-\frac{f(t)}{f(0)},~~ f(t)=\cos\left(\frac{t/T+\alpha}{1+\alpha}\cdot \frac \pi 2\right),
\end{equation}
where $\alpha$ is a hyperparameter. We test the performance of the noise schedule under the hyperparameter, which is shown in Fig.~\ref{fig: schedule_cosine}. It can be observed that as the value of $\alpha$ increases, the images tend to contain higher levels of noise.

% \subsection{Implementation Details}

% As for the schedule settings, we set $\alpha = 8e-3$. The total training diffusion time step is set as 15 for pansharpening experiments. Additionally, the exponential moving average (EMA) ratio is set to 0.995.
% % The total training iterations for different datasets are shown in Tab.. \ref{tab: iteration}. 
% Moreover, the total training iterations for WV3, GF2, and QB datasets are set to 100k, 100k, and 100k iterations, respectively. The number of sampling steps is set to 15 for all tasks.

\subsection{Datasets, Metrics and Training Details}
To demonstrate the efficacy of ResPanDiff, we conduct a series of experiments over a standard pansharpening data-collection (available at PanCollection datasett\footnote{\url{https://liangjiandeng.github.io/PanCollection.html}}). This dataset includes images from WorldView-3 (WV3, 8 bands), GaoFen-2 (GF2, 4 bands) and QB (4 bands). To comprehensively evaluate the model's performance, we carry out experiments at both reduced and full resolution. Besides, we use spectral angle mapper (SAM)~\cite{sam}, the relative dimensionless global error in synthesis (ERGAS)~\cite{ergas}, the universal image quality index (Q$2^n$)~\cite{q2n} and the spatial correlation coefficient (SCC)~\cite{scc} as fusion metrics for the reduced-resolution dataset. As for the full-resolution datasets, we employ $D_\lambda$, $D_s$, and QNR~\cite{QNR}. The proposed model method is implemented in PyTorch 2.1.1 and Python 3.10.13 using AdamW optimizer with a learning rate of $1\times 10^{-4}$ to minimize $\ell_{full}$ on an Ubuntu operating system with one NVIDIA GeForce RTX4090 GPU.

\subsection{Results}
The performance of the proposed ResPanDiff method is thoroughly evaluated across three benchmark datasets: WV3, QB, and GF2. Tables~\ref{tab: wv3_reduced_full} to \ref{tab: qb_reduced} provide a detailed comparison of ResPanDiff against a range of state-of-the-art methods, encompassing both traditional techniques and modern deep learning approaches.

\begin{figure*}[!ht]
    \centering
    \includegraphics[width=\linewidth]{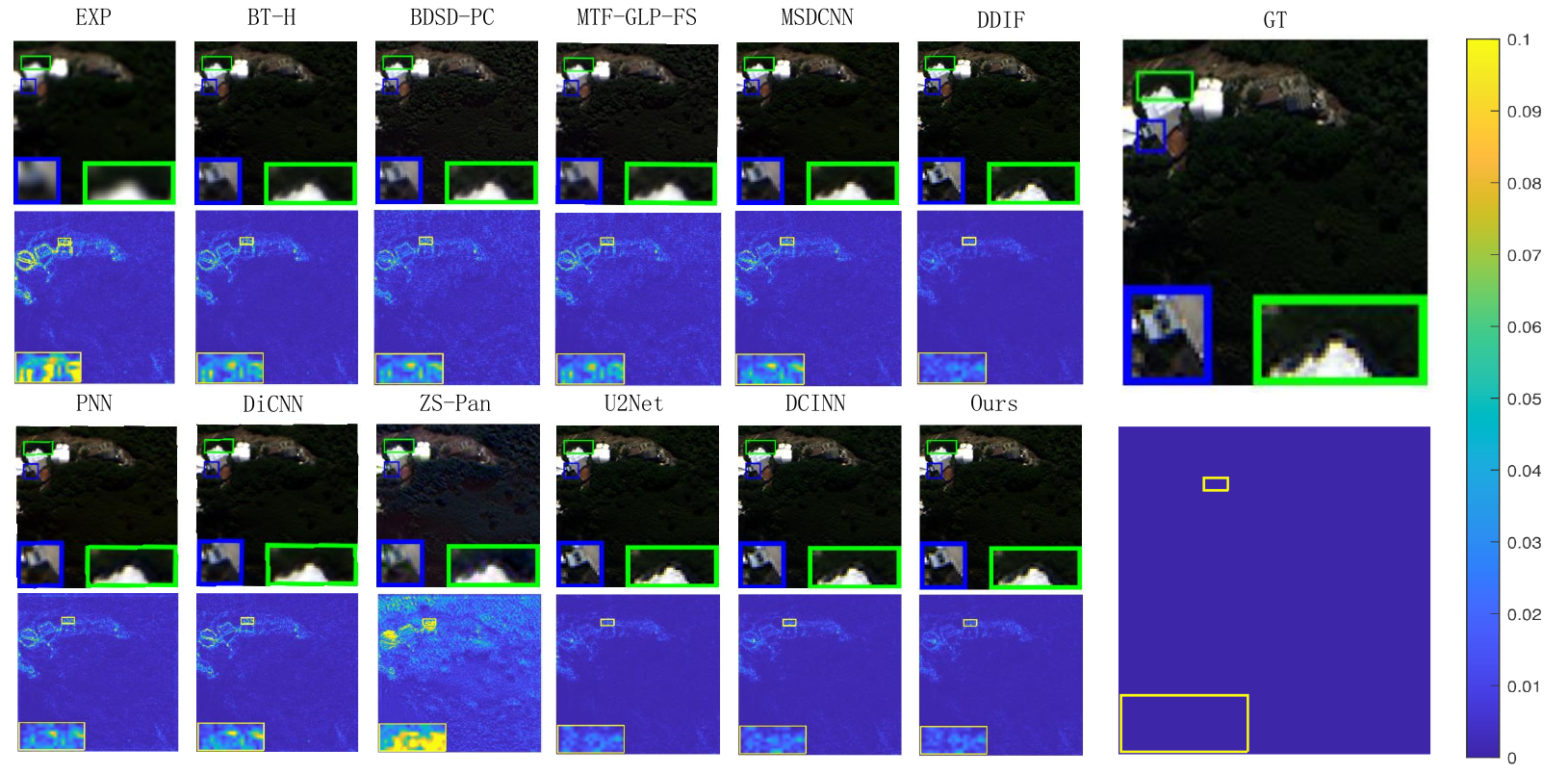}
    \caption{Visual comparisons with previous pansharpening methods on the WorldView-3 dataset. 
    The first and third rows show the RGB bands of the fused images. The second and fourth rows show the error maps between the GT and fused images by the pseudo-color. Some close-ups are depicted in red and yellow rectangles. Deeper color in the error map means better performance.
    }
    \label{fig:wv3_comp}
\end{figure*}

% \begin{figure}[t]
% \centering  %图片全局居中
% \subfigure[image 1]{
% \label{Fig.sub.1}
% \includegraphics[width=4.2cm,height = 2.8cm]{figs/exp/visual result/GT_zoom.pdf}}\subfigure[image 2]{
% \label{Fig.sub.2}
% \includegraphics[width=4.2cm,height = 2.8cm]{figs/exp/visual result/GT_error.pdf}}
% \caption{Insert two pictures side by side}
% \label{1}
% \end{figure}

\newcommand{\best}[1]{{\color{red} \textbf{#1}}}
\newcommand{\second}[1]{{\color{blue} \textbf{#1}}}
\begin{table*}[!ht]
    \centering 
    \caption{Result benchmark on the WV3 dataset, evaluated with 20 reduced-resolution samples and 20 full-resolution samples. Best results
are in red, and second-best results are in blue.}
    \resizebox{0.8\linewidth}{!}{
    \begin{tabular}{l|cccc|ccc}
        \toprule
        \multirow{2}{*}{Methods}& \multicolumn{4}{c|}{Reduced} & \multicolumn{3}{c}{Full} \\  & SAM($\pm$ std) & ERGAS($\pm$ std) & Q4($\pm$ std) & SCC($\pm$ std) & $D_\lambda$($\pm$ std) & $D_s$($\pm$ std) & QNR($\pm$ std) \\
        \midrule
        EXP~\cite{EXP} & 5.8001$\pm$1.8811 & 7.1635$\pm$1.8792 & 0.6074$\pm$0.1033 & 0.7440$\pm$0.0288 & 0.0540$\pm$0.0427 &   0.1045$\pm$0.0360 & 0.8479$\pm$0.0618\\
        BDSD-PC~\cite{bdsd-pc}         & 5.4675$\pm$1.7185 & 4.6549$\pm$1.4667 & 0.8117$\pm$0.1063 & 0.9049$\pm$0.0419  & 0.0625$\pm$0.0235 & 0.0730$\pm$0.0356 & 0.8698$\pm$0.0531 \\  
        MTF-GLP-FS~\cite{mtf-glp-fs}      & 5.3233$\pm$1.6548 & 4.6452$\pm$1.4441 & 0.8177$\pm$0.1014 & 0.8984$\pm$0.0466 & 0.0206$\pm$0.0082	& 0.0630$\pm$0.0284 & 0.9180$\pm$0.0346 \\ 
        BT-H~\cite{BT-H}         & 4.8985$\pm$1.3028 & 4.5150$\pm$1.3315 & 0.8182$\pm$0.1019 & 0.9240$\pm$0.0243 & 0.0574$\pm$0.0232 & 0.0810$\pm$0.0374 & 0.8670$\pm$0.0540 \\
        \midrule
        PNN~\cite{pnn}             & 3.6798$\pm$0.7625 & 2.6819$\pm$0.6475 & 0.8929$\pm$0.0923 & 0.9761$\pm$0.0075 & {0.0213$\pm$0.0080}	& 0.0428$\pm$0.0147 & 0.9369$\pm$0.0212 \\
        % PanNet~\cite{yang2017pannet}          & 3.6156$\pm$0.7665 & 2.6660$\pm$0.6887 & 0.8906$\pm$0.0934 & 0.9757$\pm$0.0088 & \best{0.0165$\pm$0.0074}	& 0.0470$\pm$0.0213	& \second{0.9374$\pm$0.0271} \\
        DiCNN~\cite{dicnn}         & 3.5929$\pm$0.7623 & 2.6733$\pm$0.6627 & 0.9004$\pm$0.0871 & 0.9763$\pm$0.0072 & 0.0362$\pm$0.0111	& 0.0462$\pm$0.0175  & 0.9195$\pm$0.0258 \\
        MSDCNN~\cite{msdcnn}          & 3.7773$\pm$0.8032 & 2.7608$\pm$0.6884 & 0.8900$\pm$0.0900 & 0.9741$\pm$0.0076 & {0.0230$\pm$0.0091}	& 0.0467$\pm$0.0199	& 0.9316$\pm$0.0271 \\
        % FusionNet~\cite{deng2020detail}       & 3.3252$\pm$0.6978 & 2.4666$\pm$0.6446 & 0.9044$\pm$0.0904 & 0.9807$\pm$0.0069 & 0.0239$\pm$0.0090	& 0.0364$\pm$0.0137	&{0.9406$\pm$0.0197} \\
        % CTINN~\cite{ctinn} & 3.2523$\pm$0.6436 & 2.3936$\pm$0.5194 & 0.9056$\pm$0.0840 & 0.9826$\pm$0.0046 & 0.0550$\pm$0.0288 & 0.0679$\pm$0.0312 & 0.8815$\pm$0.0488 \\
        % % LAGNet~\cite{jin2022lagconv}          & 3.1042$\pm$0.5585 & 2.2999$\pm$0.6128 & 0.9098$\pm$0.0907 & 0.9838$\pm$0.0068 & 0.0368$\pm$0.0148	& {0.0418$\pm$0.0152}	&0.9230$\pm$0.0247 \\
        % MMNet~\cite{mmnet} & 3.0844$\pm$0.6398 & 2.3428$\pm$0.6260 & 0.9155$\pm$0.0855 & 0.9829$\pm$0.0056 & 0.0540$\pm$0.0232 &  0.0336$\pm$0.0115 & 0.9143$\pm$0.0281 \\
        % DCFNet~\cite{dcfnet}          & {3.0264$\pm$0.7397} & {2.1588$\pm$0.4563} & {0.9051$\pm$0.0881} & {0.9861$\pm$0.0038} & 0.0781$\pm$0.0812 & 0.0508$\pm$0.0342 & 0.8771$\pm$0.1005 \\ 
        ZS-Pan~\cite{cao2024zero}          & {5.3000$\pm$1.2026} & {4.4397$\pm$1.1382} & {0.7846$\pm$0.0770} & {0.9339$\pm$0.0193} & 0.0185$\pm$0.0060 & 0.0279$\pm$0.0141 & 0.9542$\pm$0.0188 \\ 
        U2Net~\cite{peng2023u2net}          & {2.8888$\pm$0.5815} & {2.1498$\pm$0.5258} & {0.9197$\pm$0.0811} & {0.9863$\pm$0.0047} & {0.0178$\pm$0.0072} & 0.0313$\pm$0.0075 & 0.9514$\pm$0.0115 \\ 
        DCINN~\cite{wang2024general}          & {2.9362$\pm$0.5232} & {2.2408$\pm$0.0061} & {0.9192$\pm$0.5232} & {0.9704$\pm$0.0560} & \best{0.0158 $\pm$0.0456} & {0.0260$\pm$0.0087} & \best{0.9578$\pm$0.0417} \\ 
        DDIF~\cite{cao2024diffusion}  & \second{2.7386$\pm$0.5080} & \second{2.0165$\pm$0.4508} & \second{0.9202$\pm$0.0824} & \second{0.9882$\pm$0.0031} & 0.0258$\pm$0.0187 & {0.0231$\pm$0.0075} & {0.9517$\pm$0.0173}  \\ \midrule
        ResPanDiff(ours)  & \best{2.7179$\pm$0.4921} & \best{2.0025$\pm$0.4446} & \best{0.9284$\pm$0.0834} & \best{0.9886$\pm$0.0035} & \second{0.0164$\pm$0.0111} & \best{0.0212$\pm$0.0191} & \second{0.9531$\pm$0.0286}  \\ \midrule
        Ideal value    & \textbf{0}   & \textbf{0}  & \textbf{1}  & \textbf{1}  & \textbf{0}   & \textbf{0}  & \textbf{1} \\
        \bottomrule
    \end{tabular}
    }
    \label{tab: wv3_reduced_full}
\end{table*}

\begin{table}[!ht]
  \centering
      \caption{Result benchmark on the GaoFen-2 dataset with 20 reduced resolution samples. red: best, blue: second best.}
    \resizebox{\linewidth}{!}{
       \begin{tabular}{l|cccc}
        \toprule
        Methods & SAM($\pm$ std) & ERGAS($\pm$ std) & Q4($\pm$ std)\\
        \midrule
        EXP~\cite{EXP} & 1.8200$\pm$0.4030 & 2.3691$\pm$0.5546 & 0.8034$\pm$0.0523\\
        BDSD-PC~\cite{bdsd-pc}  & 1.7110$\pm$0.3210 & 1.7025$\pm$0.4056 & 0.9932$\pm$0.0308\\
        MTF-GLP-FS~\cite{mtf-glp-fs}  & 1.6757$\pm$0.3457 & 1.6023$\pm$0.3545 & 0.8914$\pm$0.0256 \\
        BT-H~\cite{BT-H}      & 1.6810$\pm$0.3168 & 1.5524$\pm$0.3642 & 0.9089$\pm$0.0292 \\
        \midrule
        PNN~\cite{pnn}            & 1.0477$\pm$0.2264 & 1.0572$\pm$0.2355 & 0.9604$\pm$0.0100\\
        % PanNet~\cite{yang2017pannet}          & 0.9967$\pm$0.2119 & 0.9192$\pm$0.1906 & 0.9671$\pm$0.0099   & 0.9829$\pm$0.0035 &\second{0.0206$\pm$0.0112}	&0.0799$\pm$0.0178	&0.9011$\pm$0.0203 \\
        DiCNN~\cite{dicnn}          & 1.0525$\pm$0.2310 & 1.0812$\pm$0.2510 & 0.9594$\pm$0.0101\\
        MSDCNN~\cite{msdcnn}         & 1.0472$\pm$0.2210 & 1.0413$\pm$0.2309 & 0.9612$\pm$0.0108\\
        % FusionNet~\cite{deng2020detail}    &  0.9735$\pm$0.2117    & 0.9878$\pm$0.2222   & 0.9641$\pm$0.0093\\
        % CTINN~\cite{ctinn} & 0.8251$\pm$0.1386 & 0.6995$\pm$0.1068 & 0.9772$\pm$0.0117 \\
        % % LAGNet~\cite{jin2022lagconv}       & {0.7859$\pm$0.1478}     & {0.6869$\pm$0.1125}    & {0.9804$\pm$0.0085}\\
        % MMNet~\cite{mmnet} & 0.9929$\pm$0.1411 & 0.8117$\pm$0.1185 & 0.9690$\pm$0.0204 \\
        % DCFNet~\cite{dcfnet}          & 0.8896$\pm$0.1577     & 0.8061$\pm$0.1369   & 0.9727$\pm$0.0100\\ 
        ZS-Pan~\cite{cao2024zero}    & {2.4120$\pm$0.4696} & {2.2373$\pm$0.0221} & {0.8779$\pm$0.0384}\\ 
        U2Net~\cite{peng2023u2net}        & 0.6998$\pm$0.1328     & 0.6292$\pm$0.1165   & 0.9844$\pm$0.0054\\ 
        DCINN~\cite{wang2024general}          & {0.6894$\pm$0.1055} & {0.6134$\pm$0.0221} & \second{0.9856$\pm$0.0057}\\ 
        DDIF~\cite{cao2024diffusion}  & \second{0.6408$\pm$0.1203} & \second{0.5668$\pm$0.1010} & {0.9855$\pm$0.0078} \\ \midrule
        ResPanDiff(ours)  & \best{0.6012$\pm$0.1217} & \best{0.5255$\pm$0.1005} & \best{0.9881$\pm$0.0049} \\ \midrule
        Ideal value    & \textbf{0}   & \textbf{0}  & \textbf{1} \\
        \bottomrule
        \end{tabular}
        }
        \label{tab: gf2_reduced}
\end{table}
\begin{table}[!ht]
   \centering
   \caption{Result benchmark on the QB dataset with 20 reduced resolution samples. red: best, blue: second best.}
       \resizebox{\linewidth}{!}{
         \begin{tabular}{l|cccc}
        \toprule
        Methods & SAM($\pm$ std) & ERGAS($\pm$ std) & Q4($\pm$ std)\\
        \midrule
        EXP~\cite{EXP} &  8.4354$\pm$1.9253 & 11.8368$\pm$1.9090 & 0.5667$\pm$0.0766\\
        BDSD-PC~\cite{bdsd-pc}       & 8.2620$\pm$2.0497 & 7.5420$\pm$0.8138 & 0.8323$\pm$0.1013\\  
        MTF-GLP-FS~\cite{mtf-glp-fs}      & 8.1131$\pm$1.9553 & 7.5102$\pm$0.7926 & 0.8296$\pm$0.0905\\ 
        BT-H~\cite{BT-H}            & 7.1943$\pm$1.5523 & 7.4008$\pm$0.8378 & 0.8326$\pm$0.0880\\ \midrule
        PNN~\cite{pnn}             & 5.2054$\pm$0.9625 & 4.4722$\pm$0.3734 & 0.9180$\pm$0.0938\\
        % PanNet~\cite{yang2017pannet}          & 5.7909$\pm$1.1839 & 5.8629$\pm$0.8883 & 0.8850$\pm$0.0917 & 0.9485$\pm$0.0170 & \best{0.0410$\pm$0.0108}	&0.1137$\pm$0.0323	&0.8502$\pm$0.0390  \\
        DiCNN~\cite{dicnn}           & 5.3795$\pm$1.0266 & 5.1354$\pm$0.4876 & 0.9042$\pm$0.0942\\
        MSDCNN~\cite{msdcnn} & 5.1471$\pm$0.9342 & 4.3828$\pm$0.3400 & 0.9188$\pm$0.0966\\
        % FusionNet~\cite{deng2020detail}       & 4.9226$\pm$0.9077 & 4.1594$\pm$0.3212 & 0.9252$\pm$0.0902 \\
        % CTINN~\cite{ctinn} & 4.6583$\pm$0.7755 & 3.6969$\pm$0.2888 & 0.9320$\pm$0.0072\\
        % % LAGNet~\cite{jin2022lagconv}          & 4.5473$\pm$0.8296 & {3.8259$\pm$0.4196} & {0.9335$\pm$0.0878}\\
        % MMNet~\cite{mmnet} & 4.5568$\pm$0.7285 & 3.6669$\pm$0.3036 & 0.9337$\pm$0.0941\\
        % DCFNet~\cite{dcfnet}          & {4.5383$\pm$0.7397} & 3.8315$\pm$0.2915 & 0.9325$\pm$0.0903\\ 
        ZS-Pan~\cite{cao2024zero}          & {7.7968$\pm$0.6588} & {8.3373$\pm$1.0220} & {0.9165$\pm$0.0860}\\ 
        U2Net~\cite{peng2023u2net}          & {4.7423$\pm$0.9850} & {4.5461$\pm$1.0962} & {0.9224$\pm$0.0873}\\ 
        DCINN~\cite{wang2024general}          & {4.4246$\pm$0.2731} & {3.5729$\pm$0.0071} & {0.9367$\pm$0.0868}\\ 
        DDIF~\cite{cao2024diffusion}  & \second{4.3496$\pm$0.7313} & \second{3.5223$\pm$0.2703} & \second{0.9375$\pm$0.0904}\\ \midrule
        ResPanDiff(ours)  & \best{4.3259$\pm$0.7543} & \best{3.4937$\pm$1.3837} & \best{0.9432$\pm$0.0847}\\ \midrule
        Ideal value    & \textbf{0}   & \textbf{0}  & \textbf{1}\\
        \bottomrule
    \end{tabular}
    }
    \label{tab: qb_reduced}
\end{table}

\newcommand{\cmark}{\ding{51}}%
\newcommand{\xmark}{\ding{55}}%

\subsection{Ablation Studies}
We set up various ablation studies to validate the effectiveness of the proposed methods. The overall result is shown in Tab.~\ref{ablatio: all}.

\noindent \textbf{Diffusion algorithm}
We propose a difusion algorithm to infer residual rather than full image, which not only accelerates the sampling speed but also improves the generation quality. To prove it, we ablate the algorithm, which aims to infer full image rather than residual image by training the diffusion model with different forms. The results of this ablation experiment are presented in the first to third rows of the Tab.~\ref{ablatio: all}. We can find that by modifying the diffusion model algorithm, our approach not only reduces the number of sampling steps but also improves image quality, achieving results comparable to SAM 2.7179 with 15 steps.

\noindent \textbf{Latent State}
Latent State $x_t$ is thought to have more abundant information for the model to extract, which directly results in better performance. To validate it, we ablate the input type with $x_t$ and $e_t$. The results of this ablation experiment are presented in the third to fourth rows of the Tab.~\ref{ablatio: all}. We can find that by modifying the input type, our approach has shown significant improvement, specifically, from SAM 2.8097 to SAM 2.7179.

\noindent \textbf{Shallow Cond-Injection}
To validate the effectiveness of Shallow Cond-Injection, we conduct a comparison between models that adopt SC-I and those that do not under the same diffusion algorithm. Subsequently, we retrain the models until they cconverge. The results of this ablation experiment are presented in the fifth to sixth rows of the Tab.~\ref{ablatio: all}. We can find that by modifying the input type, our approach has shown significant improvement, specifically, from SAM 2.7654 to SAM 2.7179.

\noindent \textbf{Loss Function}
We conducted ablation experiments on GF2 to validate the effectiveness of the loss function within our framework. The quantitative results demonstrate the clear advantage of our proposed loss function. The results of this ablation experiment are presented in the seventh to ninth rows of the Tab.~\ref{ablatio: all}. We can find that by modifying the diffusion model algorithm, our approach improves image quality, achieving results comparable to SAM 2.7179.

\begin{table*}[!ht]
\centering
\caption{Quantitative comparison results of pansharpening on WV3 dataset or GF2 dataset. Best results are in red.}
\resizebox{0.6\linewidth}{!}
{
\begin{tabular}{@{}ccccc|cccc|c@{}}
\toprule
\makecell{Algorithm} & \emph{T}  & Loss & input & SC-I & SAM    & ERGAS & PSNR     & SCC & Dataset    \\ \midrule
DDPM & 500 & $\ell_{res}$ & $ x_t$ & \cmark 
& 3.0482      &  3.1348 & 38.9180 & 0.9755 & WV3 \\
DDIM15 & 15 & $\ell_{res}$ & $ x_t$ & \cmark 
& 2.7602      &  2.0544 &  39.4133 & 0.9882 & WV3  \\
ResShift & 15 & $\ell_{res}$ & $ x_t$ & \cmark 
& 2.7551 & 2.0281 & 39.3853 & 0.9881 & WV3 \\
ResPanDiff & 15 & $ \ell_{res}$ & $ e_t$ & \cmark 
& 2.8097    &  2.0481 & 39.4431 & 0.9755 & WV3  \\
ResPanDiff & 15 & $ \ell_{res}$ & $ x_t$ & \xmark
& 2.7654 & 2.0433 & 39.5134 & 0.9880 & WV3 \\
ResPanDiff & 15 & $ \ell_{res}$ & $ x_t$ & \cmark 
& \best{2.7179} & \best{2.0025} & \best{39.6655} & \best{0.9886} & WV3 \\
ResPanDiff & 15 & $ \ell_{2}$ & $ x_t$ & \cmark 
& 0.6300 & 0.5575 & 44.5529 & 0.9933 &GF2 \\
ResPanDiff & 15 & $ \ell_{1}$ & $ x_t$ & \cmark 
& {0.6149} & {0.5432} & {44.7798} & {0.9939} &GF2 \\
ResPanDiff & 15 & $ \ell_{res}$ & $ x_t$ & \cmark 
& \best{0.6012} & \best{0.5255} & \best{44.8490} & \best{0.9940} &GF2 \\

\bottomrule
\end{tabular}}
\label{ablatio: all}
\end{table*}

\section{Discussion}
In this section, we will evaluate the generalization capability of ResPanDiff by assessing its performance on the WorldView-2 (WV-2) dataset, using a model trained on the WorldView-3 (WV-3) dataset. Then, we will visualize the gradient during training with different losses. After that, we will discuss a problem about the transport trajectory.

\subsection{Generalization Ability on WorldView-2 Dataset}
To evaluate the generalization performance of ResPanDiff, we test a model trained on the WV-3 dataset against the WV-2 dataset. We also compare ResPanDiff with other DL-based approaches, as presented in Tab.~\ref{tab: wv2_reduced_full}. ResPanDiff demonstrates competitive performance, highlighting its strong generalization capabilities. This generalization ability is in line with previous works that apply diffusion models in a zero-shot manner\footnote{``Zero-shot'' refers to the capability of a trained model to perform inference on a previously unseen dataset without requiring any additional fine-tuning or re-training on that specific data.}.
% This concept highlights the model's ability to generalize effectively across different datasets or tasks. In the context of diffusion models applied to pansharpening, zero-shot learning demonstrates the robustness of the model by showing that it can maintain high performance when tested on datasets that differ from the ones used during training.}.
  % to various tasks, including image inverse problems~\cite{ddnm}, image classification~\cite{diffusion_zero_shot_cls}, semantic segmentation~\cite{diffumask}, and others.

\begin{table*}[!ht]
    \centering 
    \caption{Generalization abilities of DL-based methods are compared. The best results are in red, and the second best results are in blue.}
    \resizebox{0.8\linewidth}{!}{
    \begin{tabular}{l|cccc|ccc}
        \toprule
        \multirow{2}{*}{Methods}& \multicolumn{4}{c|}{Reduced} & \multicolumn{3}{c}{Full} \\  & SAM($\pm$ std) & ERGAS($\pm$ std) & Q4($\pm$ std) & SCC($\pm$ std) & $D_\lambda$($\pm$ std) & $D_s$($\pm$ std) & QNR($\pm$ std) \\
        \midrule
     PNN~\cite{pnn} &7.1158$\pm$1.6812& 5.6152$\pm$0.9431& 0.7619$\pm$0.0928&0.8782$\pm$0.0175& 0.1484$\pm$0.0957      & 0.0771$\pm$0.0169 & 0.7869$\pm$0.0959 \\
     % PanNet~\cite{yang2017pannet} &\second{5.4948$\pm$0.7126} & \best{4.3371$\pm$0.5197}& \best{0.8401$\pm$0.0796}& \best{0.9177$\pm$0.0081}& \second{0.0317$\pm$0.0173}      & \best{0.0191$\pm$0.0123} & \best{0.9500$\pm$0.0275} \\
     DiCNN~\cite{dicnn} &6.9216$\pm$0.7898 &6.2507$\pm$0.5745 &0.7205$\pm$0.0746 &0.8552$\pm$0.0289& 0.1412$\pm$0.0661      & 0.1023$\pm$0.0195 & 0.7700$\pm$0.0505 \\
     MSDCNN~\cite{msdcnn} &6.0064$\pm$0.6377 &4.7438$\pm$0.4939 &\second{0.8241$\pm$0.0799} &0.8972$\pm$0.0109 & 0.0589$\pm$0.0421      & \second{0.0290$\pm$0.0138} & {0.9143$\pm$0.0516} \\
     % FusionNet~\cite{deng2020detail} &6.4257$\pm$0.8602 &5.1363$\pm$0.5151 &0.7961$\pm$0.0737 & 0.8746$\pm$0.0134& {0.0519$\pm$0.0292}      & 0.0559$\pm$0.0146 & 0.8948$\pm$0.0187 \\
     CTINN~\cite{ctinn} & 6.4312$\pm$0.6219 & 4.6829$\pm$0.4713 & 0.8103$\pm$0.0891 & \second{0.9139$\pm$0.0110} & 0.1722$\pm$0.0373 & 0.0375$\pm$0.0065 & 0.7967$\pm$0.0360 \\
     % LAGNet~\cite{jin2022lagconv} &6.9545$\pm$0.4739 &5.3262$\pm$0.3185 &0.8054$\pm$0.0837 &0.9125$\pm$0.0101 & 0.1302$\pm$0.0856      & 0.0547$\pm$0.0159 & 0.8229$\pm$0.0884 \\
     MMNet~\cite{mmnet} & 6.6109$\pm$0.3209 & 5.2213$\pm$0.2133 & 0.8143$\pm$0.0790 & 0.9136$\pm$0.0201 & 0.0897$\pm$0.0340 & 0.0688$\pm$0.0209 & 0.8476$\pm$0.0569 \\
     DCFNet~\cite{dcfnet} &5.6194$\pm$0.6039 &\best{4.4887$\pm$0.3764} &\best{0.8292$\pm$0.0815} &\best{0.9154$\pm$0.0083} & 0.0649$\pm$0.0357 & 0.0700$\pm$0.0219 & 0.8690$\pm$0.0233 \\
     DDIF~\cite{cao2024diffusion} &\second{}{5.3827$\pm$0.5737} & {4.6712$\pm$0.4155} & 0.8217$\pm$0.0777 &0.8993$\pm$0.0129 & \second{0.0313$\pm$0.0376} & {0.0312$\pm$0.0111} & \second{0.9388$\pm$0.0453} \\
      \midrule
     ResPanDiff(ours) &\best{5.3758$\pm$0.5979} & \second{4.5615$\pm$0.4205} & 0.8203$\pm$0.1016 &0.9096$\pm$0.0120 & \best{0.0300$\pm$0.0184} & \best{0.0249$\pm$0.0109} & \best{0.9461$\pm$0.0273} \\
     \midrule
        Ideal value    & \textbf{0}   & \textbf{0}  & \textbf{1}  & \textbf{1}  & \textbf{0}   & \textbf{0}  & \textbf{1} \\
        \bottomrule
    \end{tabular}
    }
    \label{tab: wv2_reduced_full}
\end{table*}

\subsection{The Grad with Different Loss}
\begin{figure}[!ht]
\centering
    {\includegraphics[width=0.75\linewidth]{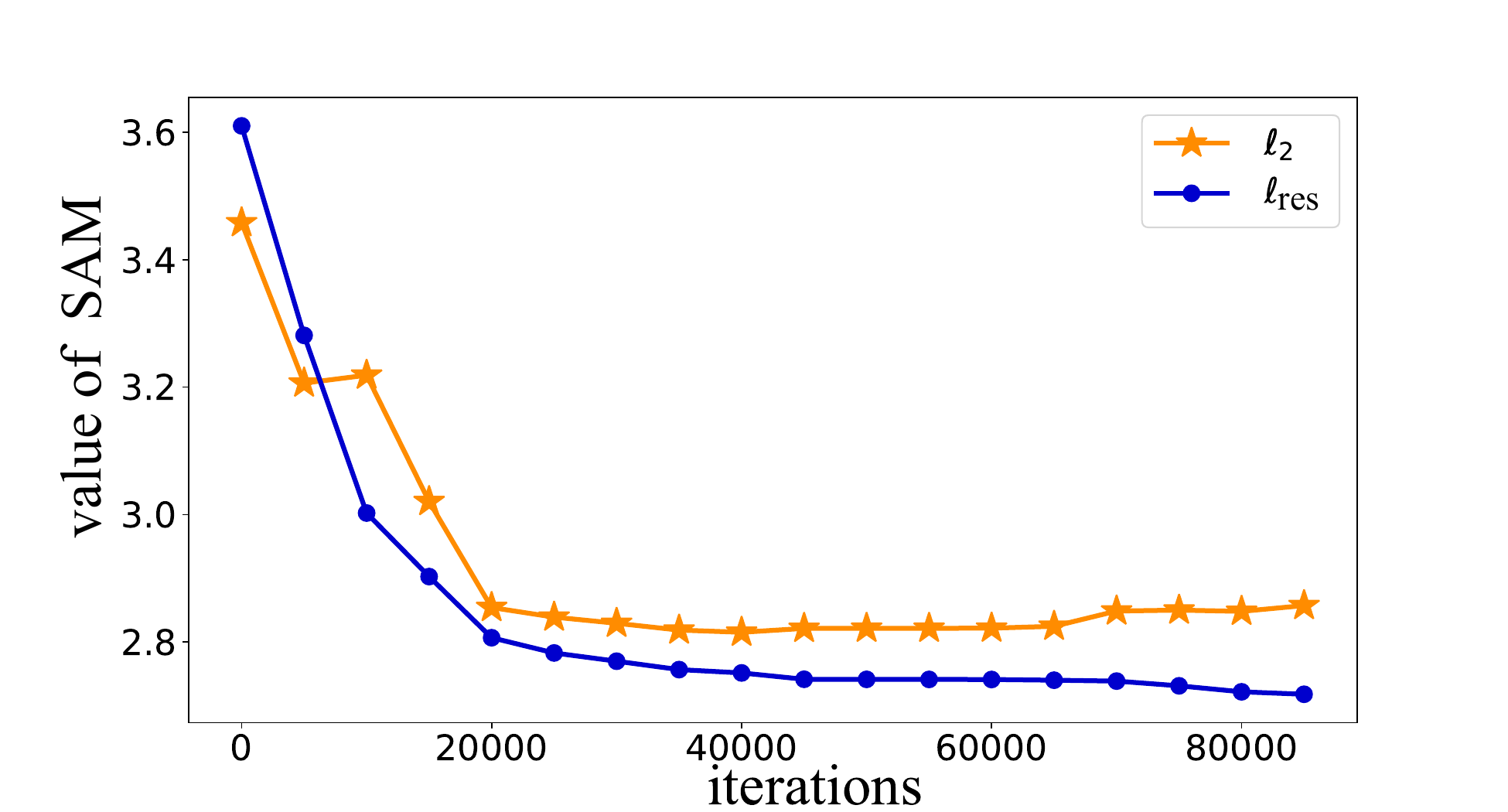}}
    \caption{SAM on different loss. We can find that at the beginning, the model utilizing $\ell_2$ loss performs better, but with the iteration increasing, the model utilizing $\ell_2$ loss becomes to fall into overfit, while the model utilizing $\ell_{res}$ continues to undergo optimization and ultimately achieves superior performance.}
    \label{fig: sam on different loss}
\end{figure}

\begin{figure}[!ht]
\centering
    {\includegraphics[width=0.75\linewidth]{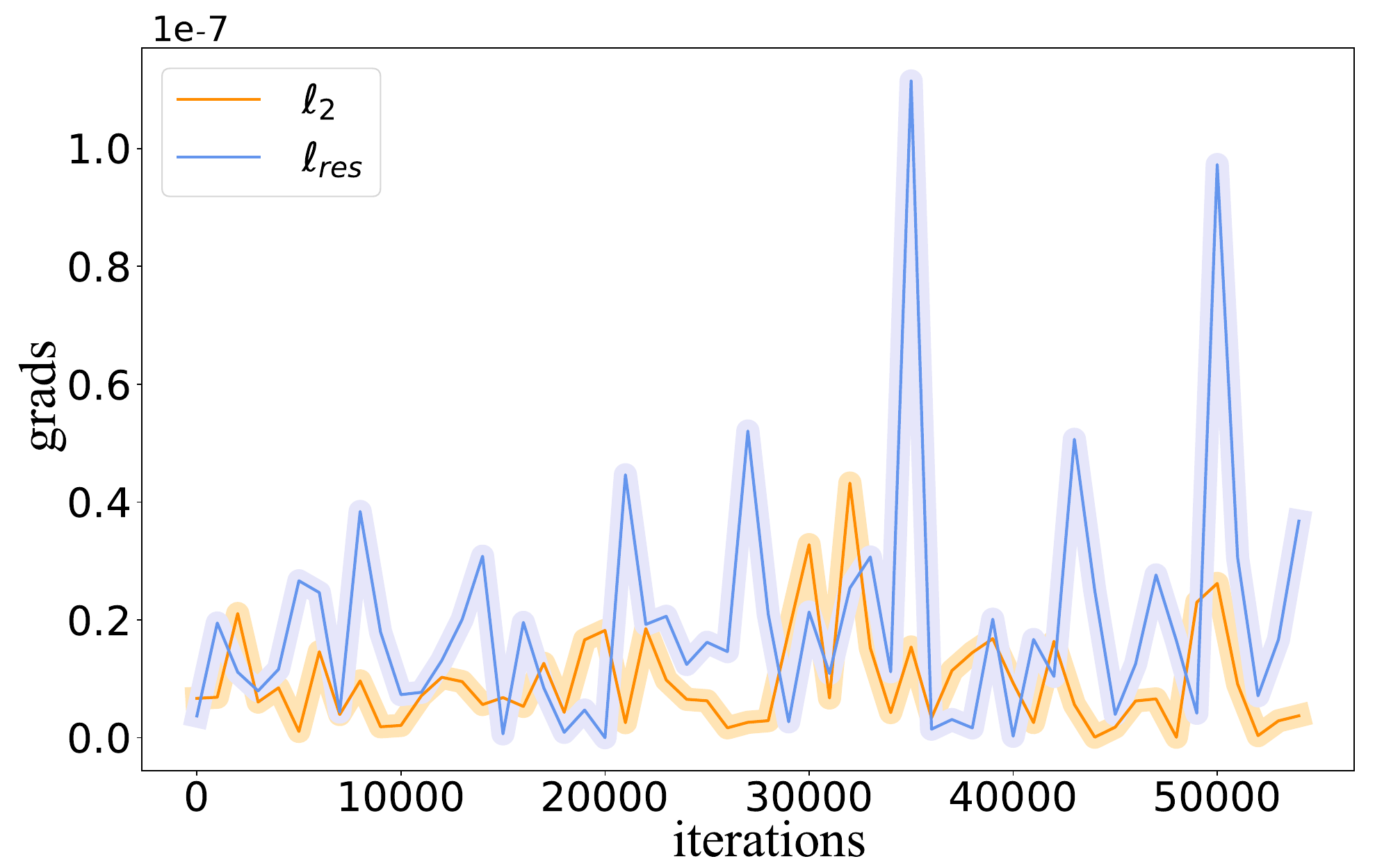}}
    \caption{Grad on different losses. We can observe that the overall gradient when using $\ell_{res}$ is greater than the gradient when using the $\ell_2$ loss function. This suggests that employing $\ell_{res}$ can provide better guidance for the model.}
    \label{fig: grad on different loss}
\end{figure}

To validate the impact of  $\ell_{res}$ function on gradients, we plot the gradients of the model's input layer and tracked the model's performance over iterations.
We can observe that, when using the $\ell_2$ loss function, the gradients are generally lower, and the model reaches its optimal performance early in the training epochs, after which it falls into overfitting, while the model ultilizing $\ell_{res}$ keeps optimizing and finally reaches better performance. This indicates that the loss function we designed effectively improves the model's performance.
The plots are shown in Fig.~\ref{fig: sam on different loss} and Fig.~\ref{fig: grad on different loss}.

\subsection{The Transport Trajectory of ResPanDiff}
\begin{figure}[H]
    \centering  %图片全局居中
    {
    \label{Fig.sub.1}
    \includegraphics[width=0.49\linewidth]{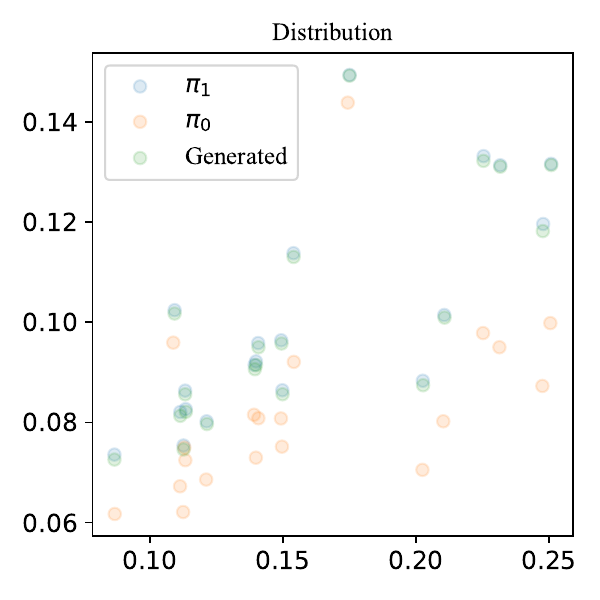}}{
    \label{Fig.sub.2}
    \includegraphics[width=0.49\linewidth]{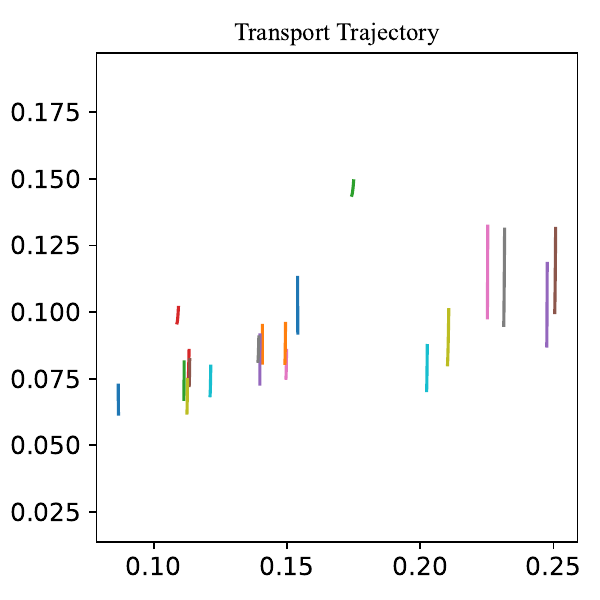}}
    \caption{The first column represents the data distribution of $LRMS \sim \pi_0$, $HRMS \sim \pi_1$ and the generated $SR$ through the generation path; the second column represents the transport trajectory from $LRMS$ to $HRMS$.}
    \label{fig: wv3 transport_test}
\end{figure}

To validate our proposed hypothesis, we plot the transport trajectory of ResPanDiff from $LRMS$ to $HRMS$ as shown in Fig.~\ref{fig: wv3 transport_test}. The experimental results align with our hypothesis, showing that the transport trajectory is relatively straight with minimal crossovers.
This also indicates that there is no need for twice training like 1-Rectified flow and 2-Rectified flow in pansharpening task. Only one time of training is enough for the model to generate the straight line. 
% \subsection{The Crossover Problems of Transport Trajectory on pansharpening Task}
% In the pansharpening task, $x_0$ and $x_T$ are correspondingly paired rather than randomly matched. Therefore, unlike image generation tasks, where crossover situations can be avoided by selecting non-intersecting paths, pansharpening does not allow us to generate alternative $x_0$ outputs by choosing other non-crossing routes. If two transport trajectory intersect, we cannot simply switch to another non-intersecting route. As a result, we must directly address the potential issue of path crossover.

% Fortunately, based on our observations of the pansharpening dataset, most transmission paths do not result in intersections. On the other hand, during the generation process, ResPanDiff receives information from $x_t$ and cond, which together provide guidance toward the target $x_0$ Therefore, for a transmission path from $x_T$ to $x_0$, we believe that even if an intersection occurs, the distribution produced by the model will be more inclined toward $x_0$ rather than any alternative HRMS image $x_0'$. This is because the presence of cond and $x_t$ directs the model toward the correct target.

% % \clearpage

%% file: sections/appendix.tex
\section{Appendix Section}
\label{sec:appendix_section}

\section{Mathematical Details} \label{appendix: math}
Before the proof, we must introduce some theroms first.
\textbf{Bayes’ theorem}:
\begin{equation}
    P(B|A)=\frac{P(AB)}{P(A)}=\frac{P(A|B)P(B)}{P(A)},
\end{equation}
where $A$ and $B$ are events and $P(A)\neq0$.

\noindent \textbf{The result of the product of Gaussian distributions}
Given two Gaussian distributions $x1\backsim\mathcal{N}(\mu_1,\sigma_1^2)$, $x2\backsim\mathcal{N}(\mu_2,\sigma_2^2)$, the result of the product of Gaussian distributions $x\backsim\mathcal{N}(\mu_,\sigma^2)$ follows:
\begin{equation}
    \begin{aligned}
    \mu &= (\sigma_1^{-2} + \sigma_2^{-2})^{-1}(\sigma_1^{-2}\mu_1 + \sigma_2^{-2}\mu_2),
     \\ \sigma^2 &= (\sigma_1^{-2} + \sigma_2^{-2})^{-1}.
    \end{aligned}
\end{equation}

\noindent \textbf{Lemma}
For $q(\boldsymbol{e}_{1:T}|\boldsymbol{e}_{0})$ defined in Eq.~\eqref{eq: p_e0|xT} and $q(e_{t-1}|e_{t},e_{0})$ in Eq.~\eqref{eq:Bayes}, we have:
\begin{equation} %\label{eq: one_forward_step}
q(e_{t}|e_{0}) \backsim {\cal N}((1-\overline{\alpha}_{t})e_{0},\kappa^2\overline{\alpha}_{t}).
\end{equation}
\emph{proof}. Assume for any $t \leq T$, $q(e_{t}|e_{0}) \backsim {\cal N}((1-\overline{\alpha}_{t})e_{0},\kappa^2\overline{\alpha}_{t}^2)$ holds, if:
\begin{equation}
    q(e_{t-1}|e_{0}) \backsim {\cal N}((1-\overline{\alpha}_{t-1})e_{0},\kappa^2\overline{\alpha}_{t-1}),
\end{equation}
then we can prove the statement with an induction argument for $t$ from $T$ to 1, since the base case $(t = T)$ already holds.
First, we have that:
\begin{equation}
    q(e_{t-1}|e_{0}) :=\int_{\boldsymbol{e}_{t}} q ( \boldsymbol{e}_{t} | \boldsymbol{e}_{0} ) q ( \boldsymbol{e}_{t-1} | \boldsymbol{e}_{t}, \boldsymbol{e}_{0} ) \mathrm{d} \boldsymbol{e}_{t},
\end{equation}
and
\begin{equation}
    q(e_{t}|e_{0}) \backsim {\cal N}((1-\overline{\alpha}_{t})e_{0},\kappa^2\overline{\alpha}_{t}),
\end{equation}
\begin{equation}
    p(e_{t-1}|e_{t},e_{0}) \sim \mathcal{N}\left(\frac{\overline{\alpha}_{t-1}}{\overline{\alpha}_t} e_{t} + \frac{\alpha_t}{\overline{\alpha}_t} e_{0}, \kappa^2 \frac{\overline{\alpha}_{t-1} }{\overline{\alpha}_t}\alpha_t  \right).
\end{equation}

we have that $p(e_{t-1}|e_{0})$ is Gaussian, denoted as $\mathcal{N}\left(\mu_{t-1}, \sigma_{t-1}\right)$ where
\begin{equation}
    \begin{aligned}
    \mu_{t-1} &= \frac{(\overline{\alpha}_{t-1}-\overline{\alpha}_{t-1}\overline{\alpha}_{t}+\alpha_t)e_0}{\overline{\alpha}_{t}}\\
    &= (1-\overline{\alpha}_{t-1})e_0,
    \end{aligned}
\end{equation}
and
\begin{equation}
    \begin{aligned}
    \sigma_{t-1}^2 &= \kappa^2 \frac{\overline{\alpha}_{t-1} }{\overline{\alpha}_t}\alpha_t + \frac{\kappa^2 \overline{\alpha}_{t-1}-\kappa^2 \frac{\overline{\alpha}_{t-1}}{\overline{\alpha}_t}\alpha_t}{\overline{\alpha}_t}\overline{\alpha}_t\\
    &= \kappa^2 \overline{\alpha}_{t-1}.
    \end{aligned}
\end{equation}
Therefore, $q(e_{t-1}|e_{0}) \backsim {\cal N}((1-\overline{\alpha}_{t-1})e_{0},\kappa^2\overline{\alpha}_{t-1})$, which allows us to apply the induction argument.

\subsection{Derivation of Eq.~\eqref{eq: one_forward_step}}
According to the transition distribution outlined in Eq.~\eqref{eq: q_one_sample}, $e_{t}$ can be sampled using the reparameterization trick described below:
\begin{equation} \label{eq: one_step_sum}
    \boldsymbol{e}_{t} = \boldsymbol{e}_{t-1} - \alpha_{t}\boldsymbol{e}_{0} + \kappa{\alpha_{t}}\mathcal{I},
\end{equation}
where $\mathcal{I} \backsim {\cal N}(0,1)$.
The forward process is specified by the approximate posterior $q(x_{1}:T | x_{0})$ as the following equation:
\begin{equation} \label{eq: iterative}
    q(\boldsymbol{e}_{1:T}|\boldsymbol{e}_{0}) = \displaystyle\prod_{t=1}^{T}q(\boldsymbol{e}_{t}|\boldsymbol{e}_{t-1}),
\end{equation}
recursively substitute Eq.~\eqref{eq: one_step_sum} into Eq.~\eqref{eq: iterative} with $t =
1, 2, \cdots, T$ , we can obtain the marginal distribution as follows:
\begin{equation} \label{eq: sum_q}
    \begin{aligned} {{{{\boldsymbol{e}_{t}}}}} & {{} {{} {{}={\boldsymbol{e}_{0}}-\sum_{i=1}^{t} \alpha_{i} {\boldsymbol{e}_{0}}+\kappa\sum_{i=1}^{t} \sqrt{\alpha_{i}}\xi_i}}} \\ {{{{}}}} & {{} {{} {{}=(1-\bar\alpha_{t} ){\boldsymbol{e}_{0}}+\kappa \sqrt{\bar{\alpha}_{t}} \xi_t,}}} \\ \end{aligned} 
\end{equation}
consequently, the marginal distribution as presented in Eq.~\eqref{eq: one_forward_step} is derived from Eq.~\eqref{eq: sum_q}.

\subsection{Derivation of Eq.~\eqref{eq:reverse}}
To begin with, Eq.~\eqref{eq: q_one_sample} can be written as:
\begin{equation} \label{eq: q1}
    q(e_{t}|e_{t-1},e_{0}) = \frac{1}{\boldsymbol{c}_1} exp[-\frac{( \boldsymbol{e}_{t-1}-(\boldsymbol{e}_{t}+\alpha_{t} \boldsymbol{e}_{0}) ) ( \boldsymbol{e}_{t-1}-(\boldsymbol{e}_{t}+\alpha_{t} \boldsymbol{e}_{0} ))^{T}} {2 \kappa^{2} \alpha_{t}} ],
\end{equation}
as is defined that:
\begin{equation}
    \begin{aligned}
    &\mu_{1} = \boldsymbol{e}_{t}+\alpha_{t}, \\
    &\Sigma_{1} = {\kappa^{} \sqrt{\alpha_{t}}} \mathcal{I},
    \end{aligned}
\end{equation}
and Eq.~\eqref{eq: one_forward_step} can be written as:
\begin{equation} \label{eq: q2}
    q(e_{t-1}|e_{0}) = \frac{1}{\boldsymbol{c}_2} exp[-\frac{( \boldsymbol{e}_{t-1}-(1-\bar{\alpha}_{t-1}) \boldsymbol{e}_{0} ) ( \boldsymbol{e}_{t-1}-(1-\bar{\alpha}_{t-1}) \boldsymbol{e}_{0})^{T}} {2 \kappa^{2} \bar{\alpha}_{t}} ],
\end{equation}
having that:
\begin{equation}
    \begin{aligned}
    &\mu_{2} = (1-\bar{\alpha}_{t-1}) \boldsymbol{e}_{0}, \\
    &\Sigma_{2} = {\kappa^{} \sqrt{\bar{\alpha}_{t-1}}} \mathcal{I},
    \end{aligned}
\end{equation}
according to Bayes' theorem, the posterior distribution can be formulated as:
\begin{equation} \label{eq: Bayes}
    p(e_{t-1}|e_{t},e_{0}) \varpropto q(e_{t}|e_{t-1},e_{0})q(e_{t-1}|e_{0}),
\end{equation}
%为了得到 的均值和方差，我们首先写出q1和q2的均值和方差：
% In order to get the mean and variance of $p(e_{t-1}|e_{t},e_{0})$, we first write out the mean and variance of $q(e_{t}|e_{t-1}$ and $e_{0})q(e_{t-1}|e_{0})$:

substituting Eq.~\eqref{eq: q1} and Eq.~\eqref{eq: q2} into Eq.~\eqref{eq: Bayes}, the mean and variance of targeted distribution $p(e_{t-1}|e_{t},e_{0})$ in Eq.~\eqref{eq: Bayes} can be processed in an explicit form given below:
\begin{equation} \label{eq: mu}
    \begin{aligned}
    \mu &= (\Sigma_1^{-2} + \Sigma_2^{-2})^{-1}(\Sigma_1^{-2}\mu_1 + \Sigma_2^{-2}\mu_2)
     \\&= \frac{\overline{\alpha}_{t-1}}{\overline{\alpha}_t} e_{t} + \frac{\alpha_t}{\overline{\alpha}_t} e_{0},
    \end{aligned}
\end{equation}
and
\begin{equation} \label{eq: var}
    \begin{aligned}
    \Sigma^2 &= (\Sigma_1^{-2} + \Sigma_2^{-2})^{-1} \\ 
    &= \kappa^2 \frac{\overline{\alpha}_{t-1} }{\overline{\alpha}_t}\alpha_t  \mathcal{I}.
    \end{aligned}
\end{equation}
Consequently, the marginal distribution as presented in Eq.~\eqref{eq:reverse} is derived from Eq.~\eqref{eq: mu}, Eq.~\eqref{eq: var}.